%% file: c2farxiv.tex
\documentclass{article}

\PassOptionsToPackage{numbers}{natbib}

\usepackage[final]{modifiedneurips}

\usepackage[utf8]{inputenc} 
\usepackage[T1]{fontenc}    
\usepackage{url}            
\usepackage{booktabs}       
\usepackage{amsfonts}       
\usepackage{nicefrac}       
\usepackage{xcolor}         

\definecolor{linkblue}{RGB}{47, 117, 187}
\usepackage[colorlinks=true,allcolors=linkblue]{hyperref}       

\usepackage{adjustbox}
\usepackage{afterpage}
\usepackage[linesnumbered,ruled,vlined,algo2e]{algorithm2e}
\usepackage{amsmath,amssymb,bm,mathtools}
\usepackage{array}
\usepackage{colortbl}
\usepackage{graphicx}
\usepackage{dsfont}
\usepackage{enumitem}
\usepackage{float}
\usepackage{hhline}
\usepackage{lato}
\usepackage{multirow}
\usepackage{setspace}
\usepackage{tablefootnote}   
\usepackage{tabularx}
\usepackage[normalem]{ulem}
\usepackage{wrapfig}
\usepackage[skip=2pt,font=footnotesize,labelfont=bf]{caption}

\newcolumntype{Y}{>{\centering\arraybackslash}X}
\newcolumntype{U}{>{\raggedleft\arraybackslash}X}

\newlength{\oldtextfloatsep}
\SetCommentSty{commentfont}

\newcommand{\code}[1]{\texttt{\small\normalfont\ttfamily\fontseries{m}\selectfont #1}}
\newcommand{\class}[1]{\texttt{\small\normalfont\ttfamily\fontseries{m}\selectfont #1}}
\newcommand{\dataset}[1]{{\lato\footnotesize #1}}
\newcommand{\method}[1]{{\lato\footnotesize #1}}
\newcommand{\model}[1]{{\normalsize\texttt{#1}}}

\definecolor{myRed}{rgb}{1.0, 0.44, 0.356}
\definecolor{myGreen}{rgb}{0.52, 0.8, 0.27}
\definecolor{myBlue}{rgb}{0.21747533000128158, 0.5305292836088684, 0.7548225041650647}

\input{coloring-table-macro}
\usepackage[nomessages]{fp}
\usepackage{tikz}
\usepackage{collcell}
\newcommand*{\MinNumber}{-4.0}%
\newcommand*{\MidNumber}{0.0} %
\newcommand*{\MaxNumber}{4.0}%

\renewcommand*\citet[1]{\cite{#1}}

\title{Coarse-to-Fine Curriculum Learning}

\author{%
  Otilia Stretcu\textsuperscript{1}, Emmanouil Antonios Platanios\textsuperscript{2}, Tom M. Mitchell\textsuperscript{1}, Barnabás Póczos\textsuperscript{1} \\
  \textsuperscript{1} Machine Learning Department, Carnegie Mellon University, Pittsburgh, USA \\
  \textsuperscript{2} Microsoft Semantic Machines, USA \\
  \texttt{ostretcu@cs.cmu.edu}
}

\begin{document}

\maketitle

\begin{abstract}
\input{abstract}
\end{abstract}

\input{intro}
\input{method}
\input{experiments}

\input{related-work}
\input{conclusion}

\section*{Acknowledgements}
This material is based upon work supported by AFOSR FA95501710218, NSF IIS1563887, DARPA/AFRL and FA87501720130. Any opinions, findings and conclusions or recommendations expressed in this material are those of the author(s) and do not necessarily reflect the views of the Air Force Office of Scientific Research, the National Science Foundation, the Defense Advanced Research Projects Agency or Air Force Research Laboratory.

\bibliography{c2farxiv}
\bibliographystyle{plain} 



\clearpage



\input{appendix.tex}

\end{document}

%% file: coloring-table-macro.tex
\newcommand{\ApplyGradient}[1]{%
        \ifdim #1 pt > \MidNumber pt
            \FPeval{\percent}{max(min(100.0*(#1 - \MidNumber)/(\MaxNumber-\MidNumber), 100.0), 0.00)}
            \hspace{-0.33em}\colorbox{myGreen!\percent!white}{#1}
        \else
            \FPeval{\percent}{max(min(100.0*(#1 - \MidNumber)/(\MinNumber-\MidNumber), 100.0), 0.00)}
            \hspace{-0.33em}\colorbox{myRed!\percent!white}{#1}
        \fi
}

\newcommand{\diff}[2]{%
    \ifdim #1 pt > \MidNumber pt
        \FPeval{\percent}{max(min(100.0*(#1 - \MidNumber)/(\MaxNumber-\MidNumber), 100.0), 0.00)}
        \xdef\percent{\percent}
        \cellcolor{myGreen!\percent!white} \phantom{-}#1 {\scriptsize $\pm$} #2
    \else
        \FPeval{\percent}{max(min(100.0*(#1 - \MidNumber)/(\MinNumber-\MidNumber), 100.0), 0.00)}
        \xdef\percent{\percent}
        \cellcolor{myRed!\percent!white} #1 {\scriptsize $\pm$} #2
    \fi
}

\newcolumntype{R}{>{\collectcell\ApplyGradient}r<{\endcollectcell}}

\newcommand{\dif}[1]{%
    \ifdim #1 pt > \MidNumber pt
        \FPeval{\percent}{max(min(100.0*(#1 - \MidNumber)/(\MaxNumber-\MidNumber), 100.0), 0.00)}
        \xdef\percent{\percent}
        \cellcolor{myGreen!\percent!white} \phantom{-}#1
    \else
        \FPeval{\percent}{max(min(100.0*(#1 - \MidNumber)/(\MinNumber-\MidNumber), 100.0), 0.00)}
        \xdef\percent{\percent}
        \cellcolor{myRed!\percent!white} #1
    \fi
}

%% file: abstract.tex
When faced with learning challenging new tasks, humans often follow sequences of steps that allow them to incrementally build up the necessary skills for performing these new tasks.
However, in machine learning, models are most often trained to solve the target tasks directly.
Inspired by human learning, we propose a novel curriculum learning approach which decomposes challenging tasks into sequences of easier intermediate goals that are used to pre-train a model before tackling the target task.
We focus on classification tasks, and design the intermediate tasks using an automatically constructed label hierarchy.
We train the model at each level of the hierarchy, from {\em coarse labels to fine labels}, transferring acquired knowledge across these levels.
For instance, the model will first learn to distinguish animals from objects, and then use this acquired knowledge when learning to classify among more fine-grained classes such as \texttt{cat}, \texttt{dog}, \texttt{car}, and \texttt{truck}.
Most existing curriculum learning algorithms for supervised learning consist of scheduling the order in which the training examples are presented to the model.
In contrast, our approach focuses on the {\em output space} of the model.
We evaluate our method on several established datasets and show significant performance gains especially on classification problems with many labels.
We also evaluate on a new synthetic dataset which allows us to study multiple aspects of our method.\looseness=-1


%% file: intro.tex
\section{Introduction}
\label{sec:introduction}

The field of artificial intelligence (AI) has witnessed an impressive leap in the last decade.
Machines are now able to perform tasks never before thought to be possible, such as driving cars or interacting with humans in natural language.
However, these advances were only possible through large data collection efforts. 
Humans, on the other hand, are very good at learning new skills efficiently using incredibly small amounts of supervision.
Inspired by this, AI researchers have often attempted to create models that resemble the way the human brain works, with notable examples being convolutional neural networks \citep{LeCun:2015:deep-learning} and various forms of attention mechanisms \citep[e.g.,][]{Vaswani:2017:attention}.
However, one key difference between human learning and machine learning (ML) that is often overlooked lies not in the model architecture, but in the way in which training examples are presented to the learner.
Unlike most ML systems, humans do not learn new difficult tasks (e.g., solving differential equations) entirely from scratch by looking at independent and identically distributed examples. 
Instead, new skills are often learned progressively, starting with easier tasks and gradually becoming able to tackle harder ones.
For example, students first learn to perform addition and multiplication before learning how to solve equations. 
Moreover, there is ample evidence in neuroscience that the human ability to understand concepts and infer relationships among them is acquired progressively.
\citet{warrington1975selective} was amongst the first to suggest that children first learn abstract conceptual distinctions, before progressing to finer ones.
Subsequent studies have also found evidence consistent with this coarse-to-fine progression \citep{mandler1992build, mandler1993concept, mandler2000perceptual, pauen2002global, mcclelland2003parallel, keil2013semantic}.
Thus, we can think of human learning as being driven by a {\em curriculum} that is either explicitly provided by a teacher, or implicitly learned. 
In this paper, we propose such a {\em coarse-to-fine curriculum} algorithm for ML systems.

\begin{figure}[t!]
    \vspace{-2ex}
    \centering
    \includegraphics[width=0.7\linewidth]{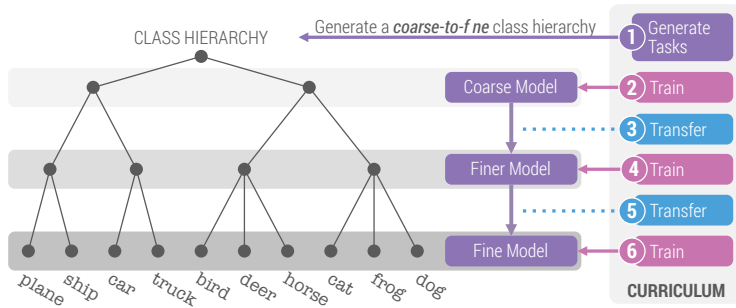}
    \caption{High-level illustration of the proposed algorithm.}
    \label{fig:overview}
    \vspace{-3ex}
\end{figure}

Using curricula in the context of machine learning was first proposed by \citet{Elman:1993:learning}, before \citet{Bengio:2009:curriculum} coined the term {\em curriculum learning}.
\citet{Bengio:2009:curriculum} and others \citep[e.g.,][]{Wang:2018:mancs,Zhou:2018:minimax,Jiang:2015:self-paced-curriculum,Jiang:2018:mentor-net} focus on scheduling the order in which training data is presented to the learner, and rely on the assumption that training datasets generally contain examples of varying {\em difficulty}.
We refer to this form of curriculum learning as {\em curriculum in input space}.
Humans also often use this form of curriculum (e.g., when learning foreign languages).
However, as mentioned earlier, human learning of semantic concepts follows a different type of curriculum.
For example, when a baby encounters a dog for the first time, her parents teach her that it is simply a ``dog,'' rather than specifying its breed.
Only later on, they start helping her distinguish between different dog breeds.
We refer to this type of curriculum as {\em curriculum in output space}. Perhaps surprisingly, this idea has been underexplored in ML. 

\begin{wrapfigure}{r}{0.35\linewidth}
    \centering
    \vspace{-5.5ex}
    \includegraphics[width=\linewidth]{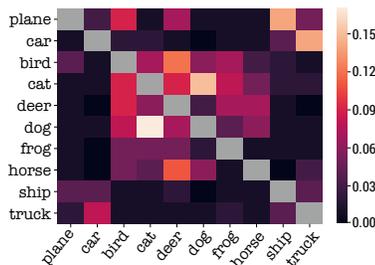}
    \caption{
        Confusion matrix for a CNN on \dataset{\footnotesize CIFAR-10}.
        Position $(i,j)$ indicates the ratio of times the correct class $i$ is confused for class $j$.
        The diagonal was removed to make small differences among the other elements discernible.\looseness=-1}
    \label{fig:confusion}
    \vspace{-2ex}
\end{wrapfigure}

Aside from the human inspiration, there are also other
reasons why a curriculum in output space would be desirable. 
Curriculum in input space is appropriate for cases where we can rank the training examples by difficulty (e.g., as \citet{Platanios:2019:competence} mention, short sentences are easier to translate than longer ones).
However, we argue that for some types of problems the errors that a model makes can be attributed to a large extent to similarities in the output space.
For example, in classification tasks it is often the case that a model gets confused due to the {\em similarity of the classes} being considered. 
To exemplify this, in Figure~\ref{fig:confusion} we present the confusion matrix of a convolutional neural network classifier (for details see Section~\ref{sec:experiments}) trained on the popular \dataset{CIFAR-10} dataset~\cite{Krizhevsky:2009:cifar-10}, which shows that the errors are not uniformly distributed across all pairs of classes.
Instead, they are mostly dominated by a select few class pairs that are difficult to distinguish between (e.g., \class{dog} and \class{cat}).
Note also that certain classes like \class{car} are mainly confused with only one or a few other classes, suggesting that an image is often difficult to classify correctly because of its similarity to images that correspond to a few specific other classes. 
Another potential disadvantage of input space curriculum approaches is that training datasets usually contain only examples for the difficult target task (e.g., distinguishing between multiple different species of animals), but no examples at all for easy intermediate goals (e.g., distinguishing between mammals and reptiles).
For example, this is true for ImageNet~\citep{Russakovsky:2015:imagenet}, a popular image classification dataset.
In these situations, we would ideally like for a system to be able to break a difficult learning task down into a sequence of easier sub-tasks that better facilitate learning.
This motivates the design of curricula that operate on the learning tasks themselves, rather than on the order in which training data is presented.\looseness=-1 


To this end, we propose a novel algorithm for performing curriculum learning on the {\em output space} of a model.
This algorithm is aimed at classification problems and enables learners to decompose difficult problems into sequences of {\em coarse-to-fine} classification problems, that improve learning for the original difficult problem.
Our main goal is to answer the following questions:
\begin{itemize}[noitemsep,label=--,leftmargin=2em,topsep=0pt]
    \item How can we automatically construct a sequence of learning tasks from {\em coarse}- to {\em fine}-grained?
    \item How can knowledge acquired by learning coarse-grained tasks transfer to fine-grained tasks?
    \item How does such a curriculum learning approach affect the generalization ability of a model?
\end{itemize}
The proposed method allows us to answer these questions in a model-independent manner, and can be applied to any classification problem without requiring additional human supervision.
We also perform an empirical evaluation 
using several established classification datasets and show that the proposed method is able to consistently boost the performance of multiple baseline models.

%% file: method.tex
\section{Proposed Method}
\label{sec:method}

Consider a classification task that involves $K$ mutually-exclusive classes.
Given a dataset of supervised examples, $\{x_i, y_i\}_{i=1}^N$, our goal is to learn a classification function $f_{\theta}: \mathcal{X} \to \mathcal{Y}$ that is parameterized by $\theta$. 
$\mathcal{X}$ can refer to an arbitrary input domain (e.g., images, text sentences, etc.) and $\mathcal{Y} = \mathds{1}^K$ contains the one-hot encoded representation of the target class for each sample.
In what follows, we represent the set of training examples $\{x_i, y_i\}_{i=1}^N$ as two tensors, $\bm{X}$ and $\bm{Y}$, that contain all of the training examples stacked along the first dimension.
The standard strategy for learning $f_\theta$ is to initialize $\theta$ using random values and iteratively update it by performing gradient descent on a loss function that is defined over $\bm{X}$ and $\bm{Y}$.
In this work, we propose a different approach: we learn a series of auxiliary functions $f_{\theta_1}, f_{\theta_2}, \hdots, f_{\theta_M}$, sequentially, where the final function $f_{\theta_M}$ corresponds to our target function, $f_\theta$.
These functions operate on the same input domain as $f_{\theta}$, but the task they are learning is coarser, meaning that they each learn to classify samples into fewer classes than the function that comes after them.
This means that $f_{\theta_1}$ is learning an easier task than $f_{\theta_2}$, $f_{\theta_2}$ an easier task than $f_{\theta_3}$, etc., up until $f_{\theta_M}$, which is learning our actual target task.
Our method thus consists of two parts:
(i) deciding what the auxiliary tasks should be and providing a way to automatically generate them along with training data for them, and
(ii) providing a way for each learned function to transfer its acquired knowledge to the next function in the chain.
An illustration of the proposed approach is shown in Figure~\ref{fig:overview}.

\subsection{Generating Auxiliary Tasks}
\label{sec:auxiliary-tasks}

Our main requirement for the auxiliary learning tasks is that they form a sequence of increasing difficulty.
We posit that grouping similar classes into coarse clusters will lead to an easier classification task. But how can we automatically decide which classes are similar?

\paragraph{Measuring Class Similarity.}
There exists a natural heuristic for gauging how similar classes are, and that is the confusion matrix of a trained classification model.
However, it turns out that using this as our class similarity metric results in a degenerate case that we discuss in Appendix~\ref{sec:confusion-degenerate-case}. 
Thus, we consider another similarity metric that also encodes what the trained model might find confusing: the {\em class embedding similarity}.
We define the embedding of a target class as the parameters of the final layer of the trained model, that are associated with that class.
In a neural network setting, the final layer typically consists of a linear projection using a weight matrix $\bm{W} \in \mathbb{R}^{E \times K}$ that maps from the last hidden layer of size $E$ to the predicted logits, for each of the $K$ classes.
We use $\bm{W}_{\cdot k}$, the $k$-th column of $\bm{W}$, to represent the embedding of class $k$.
Thus, we can measure the distance between two classes, $k_1$ and $k_2$, as $d(k_1, k_2) = \cos(\bm{W}_{\cdot k_1}, \bm{W}_{\cdot k_2})$, where $\cos$ refers to the cosine distance, and their similarity as $1 - d(k_1, k_2)$.

\paragraph{Defining a Coarse Classification Task.}
Given the original set of classes and their computed similarities, we expect that:
(i) grouping together similar classes to form coarse clusters, and
(ii) defining a new classification task where the goal is to predict the cluster instead of the specific class,
should result in an easier learning problem.
Using an example from the \dataset{CIFAR-100} dataset, we could group the classes \class{willow\_tree}, \class{oak\_tree} and \class{pine\_tree} into a single \class{tree} cluster, and all samples that belong to either of these three classes receive a new label associated with this cluster.
The clusters allow us to define a new {\em coarser} classification problem that the auxiliary function $f_{\theta_{M-1}}$ is responsible for learning.
Note that it is easy to automatically generate training data for the new task given the data of the original task:
for every training example $(x_i, y_i)$ in the original dataset, we replace the label $y_i$ with the index of its cluster.
We repeat this process for $f_{\theta_{M-2}}$,..., $f_{\theta_{1}}$.

\paragraph{Defining Coarse-to-Fine Task Sequences.}
It remains to show how we generate {\em sequences} of such auxiliary tasks with increasing difficulty, leading to the original task.
Extending the previous idea of clustering classes, we consider a {\em hierarchical clustering} algorithm.
An example of a class hierarchy is shown in Figure~\ref{fig:overview}.
If such a hierarchy forms a tree, then we can consider each level of the tree as a separate task, and form a sequence of tasks by iterating over these levels in top-down order.
In Figure~\ref{fig:overview}, the first auxiliary task, which corresponds to $f_{\theta_{1}}$ is a binary classification problem where the $2$ classes correspond to the clusters \{\class{airplane}, \class{ship}, \class{car}, \class{truck}\} and \{\class{bird}, \class{deer}, \class{horse}, \class{cat}, \class{frog}, \class{dog}\}.
The subsequent task, which corresponds to $f_{\theta_2}$, further splits these clusters resulting in $4$ classes.
Intuitively, we expect the auxiliary tasks built in this manner to be sorted by difficulty.
In other words, $f_{\theta_1}$ should be easier than $f_{\theta_2}$, $f_{\theta_2}$ than $f_{\theta_3}$, etc.
This is because, using our example from Figure~\ref{fig:overview}, $f_{\theta_3}$ would need to be able to tell whether a sample is a \class{car} or a \class{truck}, as opposed to $f_{\theta_2}$ which only needs to be able to tell if it is a \class{road vehicle}.
Formally, we have a cluster hierarchy of depth $M$ where the bottom level corresponds to the original classes.
Training data for each of these tasks can be generated automatically using the approach described in the previous paragraph.
The concrete algorithm is shown in Algorithm~\ref{alg:transform-targets}.
These tasks will be trained in order, starting with the top level in the tree, and transferring acquired knowledge from each level to the next, using the approach described in Section~\ref{sec:knowledge-transfer}.

\input{algorithms/algo-transform-labels}

\paragraph{Generating Class Hierarchies.} 
We still need to show how to automatically generate our hierarchies without human supervision.
Following our intuition from earlier in this section, at each level of the hierarchy we want to group together the most easily confused classes of the next level.
We already mentioned that we would like to use a hierarchical clustering algorithm using the similarity metric defined earlier. 
Most existing hierarchical clustering algorithms \cite{Sibson:1973:slink,Defays:1977:efficient,Kaufman:2009:finding-groups-in-data} do not directly fit our setting because they typically output binary trees, where each non-leaf node has exactly two children clusters in the next level (e.g.,  \class{cat}, \class{frog}, \class{dog} would not be directly grouped together in Figure~\ref{fig:overview}).
This can be important if we want our curriculum to visit all levels because, in the worst case, the depth of generated hierarchies will be $\mathcal{O}(K)$, where $K$ is the original number of classes.
To address this, we adopt the {\em affinity clustering} algorithm proposed by \citet{Bateni:2017:affinity}, which is based on Bor\r{u}vka's algorithm for minimum spanning trees.
This algorithm has several desirable properties---including the fact that it is parallelizable---but the property that is most important for our approach is that the depth of the hierarchy is at most $\mathcal{O}(\log K)$, where $K$ is the original number of classes.
In summary, for every level $l$ in the hierarchy, affinity clustering starts with the clusters from level $l+1$ and then joins each cluster with the one closest to it from the same level, thus forming a larger cluster.
This means that in each level
the size of the smallest cluster at least doubles relative to the next level. 
An overview of our algorithm for generating class hierarchies is shown in Algorithm~\ref{alg:generating-class-hierarchy}.

\input{algorithms/algo-gen-hierarchy}

\subsection{Transferring Acquired Knowledge}
\label{sec:knowledge-transfer}

A direct approach to transferring knowledge from a trained classifier at one level of the hierarchy to the next is via the model parameters.
We can initialize the parameters of $f_{\theta_{l+1}}$ based on the parameters of the trained $f_{\theta_l}$, for $l \in \{1,\hdots,M-1\}$.
However, since $f_{\theta_{l+1}}$ and $f_{\theta_l}$ make predictions for different number of classes, the number of parameters in $\theta_{l+1}$ does not directly match that of $\theta_{l}$.
One solution is to transfer only the subset of parameters that $\theta_{l+1}$ and $\theta_{l}$ can have in common (e.g., all except for the very last layer), and train from scratch the final prediction layer at each level of the hierarchy. 
We have attempted this approach and discuss it in detail in Appendix~\ref{app:staged} (we refer to this as the {\em staged} variant of our curriculum learning algorithm).
However, its main disadvantage is that the prediction layer, being re-initialized at each hierarchy level, can potentially lose valuable information \citep[this is often called representational collapse in pre-training literature; e.g.,][]{aghajanyan2020better}.
Ideally, we would like to be able to reuse the knowledge captured by the prediction layer used for the coarse labels when initializing the prediction layer for finer labels.
We can achieve this using the following strategy.\looseness=-1

Let us assume that we use the same model for all stages, and that it is a model that predicts the probability of each class from our target task, while not being aware of the cluster hierarchy.
When training the model at hierarchy level $\ell$, we want to use that level's cluster assignments as the target labels (instead of the original classes), and we need to define a way to supervise the model with that information.
Intuitively, when we are told that the label for an example is cluster $k$, we know that the underlying class is one that belongs to cluster $k$, but we do not know which one.
Therefore, while we are doing maximum likelihood optimization during training (e.g., by minimizing the cross-entropy function), we propose to marginalize out the class variable which is unobserved.
Given that all classes are mutually exclusive, this results in the following objective for $f_{\theta_\ell}$ (i.e., the negative log-likelihood):\looseness=-1
\begin{equation}
    \mathcal{L}_\ell = - \sum_{i} \log \!\! \sum_{c \in \mathcal{C}_\ell(y_i)} \!\! \exp \{ f_{\theta_\ell}(x_i) \},
\end{equation}
where $\mathcal{C}_\ell(y_i)$ is the cluster in level $\ell$ that class $y_i$ belongs to.
Using this formulation, the coarse-to-fine algorithm proceeds as follows (shown in  Algorithm~\ref{alg:coarse-to-fine-curriculum-continuous}):
\begin{enumerate} 
    \item We start by initializing the parameters $\theta_1$ randomly.
    \item We learn $f_{\theta_1}$ using $\mathcal{L}_1$ as the loss function.
    \item We initialize $\theta_2 = \theta_1$, and continue training using $\mathcal{L}_2$, to learn $f_{\theta_2}$.
    We can do this because the function being learned is the same for all levels.
    \item We iterate over this process until we go through all the levels of the hierarchy.
    That is, for each level $\ell$, we initialize $\theta_\ell = \theta_{\ell-1}$ and learn $f_{\theta_\ell}$ by optimizing $\mathcal{L}_\ell$.
\end{enumerate}
This allows us to learn a single function $f_{\theta}$ by going through the levels of our class hierarchy sequentially and adjusting the objective function we are optimizing appropriately.

\input{algorithms/algo-curriculum-continuous}

\subsection{Algorithm Properties}
\label{sec:method-analysis}

\paragraph{Hyperparameters.}
The proposed approach introduces a single hyperparameter: the total number of epochs to be spent on the curriculum before training on the original classes (i.e., the last level in the cluster hierarchy), which we denote as $T$.
We split this budget of $T$ epochs equally among the levels of the hierarchy.
In our experiments, we found that $T$ is easy to set using a heuristic that, albeit not optimal, consistently results in an accuracy boost across all datasets:
following \citet{Platanios:2019:competence}, we set $T$ to the number of epochs it takes for the baseline model to reach $90\%$ of its best validation accuracy.
$T$ set this way tends to be only a small fraction of the total number of training epochs ($\sim$5-10\%).\looseness=-1

\paragraph{Computational Complexity.}
Let $\mathcal{C}$ be the computational complexity required to train the baseline model to convergence.
The computational cost per training iteration of our coarse-to-fine model is approximately the same as that of the original model (label marginalization is implemented efficiently as a matrix multiplication).
Also, even though a computational overhead could come from the need to train the baseline model first in order obtain the class similarity matrix, if one is not provided, in our experiments we observed that this can be avoided.
Specifically, we observed that the relative class similarities were mostly consistent among different models (e.g., small CNN, WideResnet, Resnet), even when training on only a subset of training examples for a small number of epochs.\looseness=-1


\paragraph{Human Supervision.}
Our algorithm does not require supervision for deciding on the class hierarchy or other measures of data difficulty.
This is in contrast to many existing curriculum methods \citep[e.g.,][]{Bengio:2009:curriculum,spitkovsky2010baby,Platanios:2019:competence}.
However, prior knowledge can still be incorporated into our method by either replacing the task generation module with a provided hierarchy, or by providing a custom class dissimilarity matrix to the hierarchical clustering algorithm.
We consider this flexibility an advantage of our approach.\looseness=-1


\paragraph{Relationship to Hierarchical Classification.}
The algorithm proposed in this paper bears some conceptual similarities to hierarchical classification.
They both leverage a label hierarchy in order to obtain a better classifier.
However, there is one fundamental distinction between hierarchical classification and curriculum learning more generally: hierarchical classification methods typically use the class hierarchy, not just during training, but also while making predictions at inference time (see Section~\ref{sec:related-work} for more details).
On the other hand, curriculum learning is aimed at providing a better means of training a model, without introducing additional memory or computational costs when the model is deployed.
We propose a curriculum learning approach.
Nevertheless, it could be converted to a hierarchical classification method by creating an ensemble using the classifiers trained at each level, but this is beyond the scope of this paper.
We discuss this more extensively in Section~\ref{sec:related-work}.\looseness=-1

%% file: algorithms/algo-transform-labels.tex
{
\centering
\setlength{\textfloatsep}{10pt}
\begin{algorithm2e}[h]
\setstretch{1.1}
\caption{Transform Labels}
\label{alg:transform-targets}
\footnotesize
\tcp{\code{\scriptsize This algorithm replaces the original sample labels with their corresponding cluster index.}}
\SetKwInOut{Input}{Inputs}
\Input{Original labels $\smash{\{y_i\}_{i=1}^N}$.\\
Set of clusters $\smash{\{c_{\hat{k}}\}_{\hat{k}=1}^{\hat{K}}}$, where \\ each cluster $c_{\hat{k}}$ is a set of labels.}
$\code{originalToNew} \leftarrow$ Zero-initialized array of length K.\\
\For{$\hat{k} \leftarrow 1,\hdots,\hat{K}$}{
    \ForEach{{\normalfont Label }$l \in c_{\hat{k}}$}{
        $\code{originalToNew[}l\code{]} \leftarrow \hat{k}$
    }
}
$\code{newLabels} \leftarrow$ Zero-initialized array of length $N$.\\
\For{$i \leftarrow 1,\hdots,N$}{
    $\code{newLabels[}i\code{]} \leftarrow \code{originalToNew[}y_i\code{]}$
}
\KwOut{$\code{newLabels}$.}
\afterpage{\global\setlength{\textfloatsep}{\oldtextfloatsep}}
\end{algorithm2e}
\par
}

%% file: algorithms/algo-gen-hierarchy.tex
{
\centering
 \vspace{-1ex}
\begin{algorithm2e}[h]
\setstretch{1.1}
\caption{Generate Class Hierarchy}
\label{alg:generating-class-hierarchy}
\footnotesize
\tcp{\code{\scriptsize This algorithm generates a class hierarchy.}}
\SetKwInOut{Input}{Inputs}
\Input{Number of classes $K$.\\
Training data $\{x_i, y_i\}_{i=1}^N$.\\
Trained baseline model $f_{\theta}$.}
Estimate class distance matrix $\mathbf{D}$ by computing pairwise cosine distances between the columns of the projection matrix in $\theta^\texttt{pred}$. \\
Compute the class hierarchy, $\mathcal{H}$, using affinity clustering with $\mathbf{D}$ distance matrix between samples.\\
$\code{clustersPerLevel} \leftarrow \code{[]}$\\
\For{$l \leftarrow 1,\hdots,\code{depth}(\mathcal{H})$}{
    $\code{clustersPerLevel[}l\code{]} \leftarrow \code{[]}$\\
    \ForEach{$n \in \mathcal{H}\code{.nodesAtDepth[}l\code{]}$}{
        Create cluster $c$ by grouping the leaves of the sub-tree rooted at $n$.\\
        $\code{clustersPerLevel[}l\code{].append(}c\code{)}$\\
    }
}
\KwOut{$\code{clustersPerLevel}$.}
\end{algorithm2e}
\par
}

%% file: algorithms/algo-curriculum-continuous.tex
{
\centering
	\begin{algorithm2e}[t!]
	\setstretch{1.1}
	\caption{Coarse-To-Fine Curriculum Learning}
	\label{alg:coarse-to-fine-curriculum-continuous}
	\footnotesize
	\tcp{\code{\scriptsize This is an overview of the proposed continuous curriculum algorithm.}}
	\SetKwInOut{Input}{Inputs}
	\Input{Number of classes $K$.\\
	Training data $\{x_i, y_i\}_{i=1}^N$.\\
	Trainable baseline model $f_{\theta}$.}
	Train $f_{\theta}$ on the provided training data $\{x_i, y_i\}_{i=1}^N$. \\
    $\code{bestEpoch} \leftarrow $ epoch where $f_{\theta}$ reached its best validation accuracy \\
    \tcp{\code{\scriptsize Assign the number of epochs to spend on the curriculum either manually,  or automatically based on the baseline accuracy per epoch.}}
    $ \code{numEpochsCurriculum} \leftarrow \code{round}(\code{bestEpoch} * 0.9)$ \\
	$\code{clustersPerLevel} \leftarrow \code{\textcolor{myBlue}{GenerateClassHierarchy}(}$ $K, \{x_i, y_i\}_{i=1}^N, f_{\theta}\code{)}$\\
	$\code{M} \leftarrow  \code{clustersPerLevel.length} $\\ 
	$\code{numEpochsPerLevel} \leftarrow \code{round(numEpochsCurriculum / M)}  $\\ 
	\tcp{Train the model at each level of the hierarchy.}
	$\code{originalLabels} \leftarrow \code{[1,...,K]}$\\
	$\theta_0 \leftarrow \code{random()}$ \\
	\For{$l \leftarrow 0,\hdots,\code{M - 1}$}{
	    $\code{clusters} \leftarrow \code{clustersPerLevel[}l + 1\code{]}$\\
	    $\code{newLabels} \leftarrow \code{\textcolor{myBlue}{TransformLabels}(}$ $\{y_i\}_{i=1}^N\code{, clusters)}$\\
	    Train $f_{\theta_{l+1}}$ using $\code{newLabels}$ as the target labels, and summing the predicted probabilities for all labels in the same cluster to obtain the cluster probability, when computing the loss function. Please refer to our code repository for implementation details (e.g., how to make this calculation numerically stable).
	}
	\KwOut{$f_{\theta_{\code{[\code{M}]}}}$.}
	\afterpage{\global\setlength{\textfloatsep}{\oldtextfloatsep}}
	\end{algorithm2e}
\par
}

%% file: experiments.tex
\vspace{-1ex}
\section{Experiments}
\label{sec:experiments}
\vspace{-1ex}


We performed experiments on both synthetic and real datasets, using multiple different neural network architectures: a convolutional neural network with 3 convolution layers followed by a single densely connected layer (which we refer to as \model{CNN}), as well as the common larger models \model{Resnet-18}, \model{Resnet-50} \citep{he2016deep} and \model{WideResnet-28-10} \citep{Zagoruyko2016WideRN}.
Details on the hyperparameters we used and our training pipeline can be found in Appendix~\ref{app:exp-details}.

\begin{figure}
\vspace{-1ex}
\centering
\hfill
\begin{minipage}{.35\textwidth}
  \centering
  \includegraphics[width=0.75\textwidth]{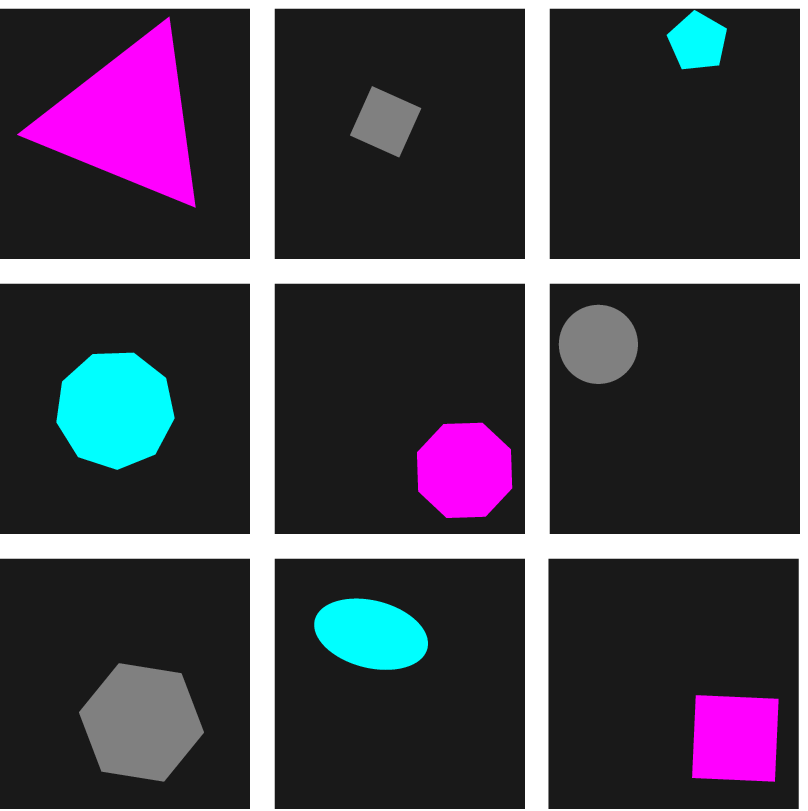}
  \vspace{0.5em}
  \captionof{figure}{Example images from the \dataset{Shapes} dataset.}
  \label{fig:shapes}
\end{minipage}%
\hfill
\begin{minipage}{.55\textwidth}
  \centering
  \includegraphics[width=0.8\textwidth]{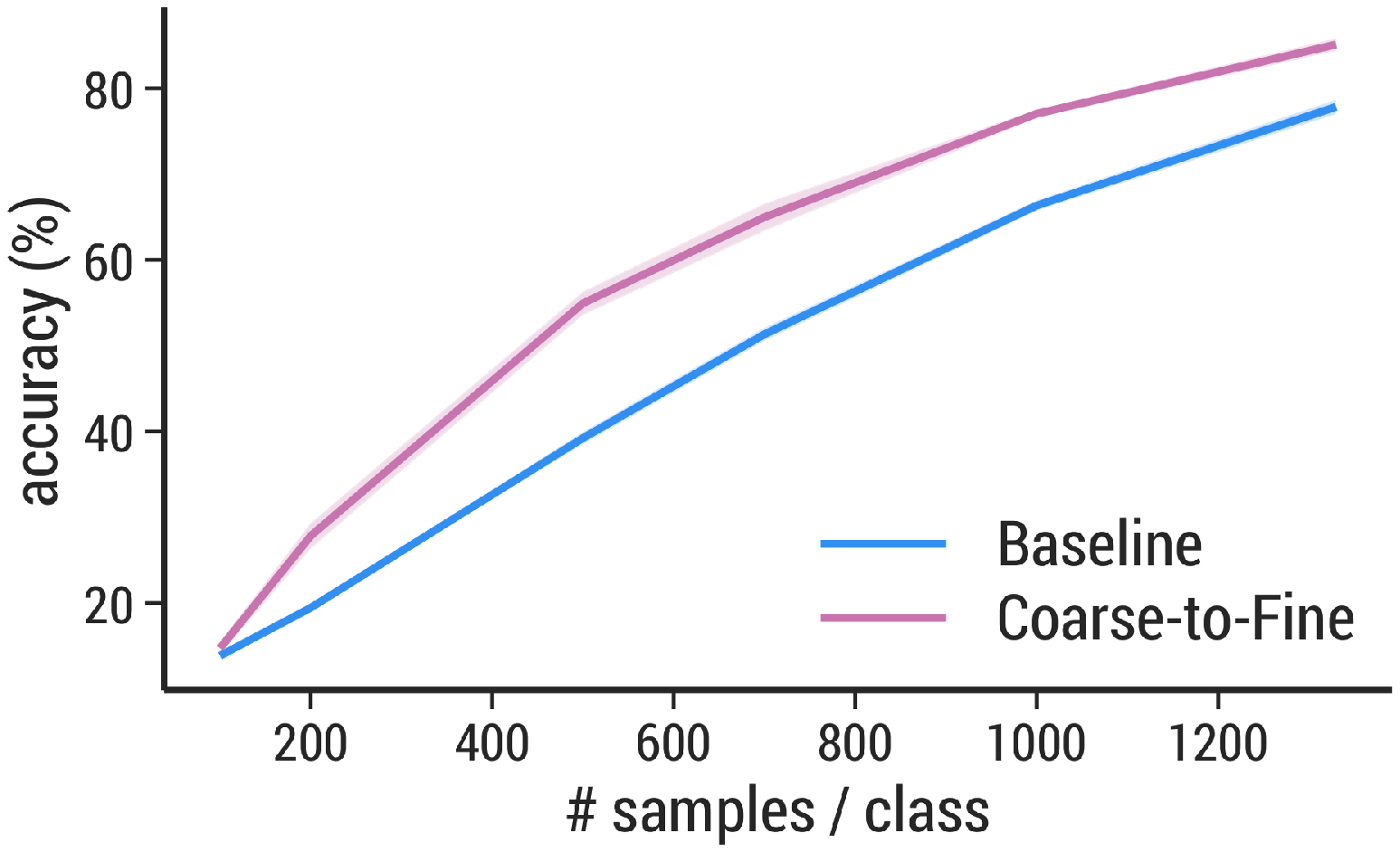}
  \captionof{figure}{Accuracy mean and standard error (computed over 5 runs and shown in tight bands around the mean) for a \model{CNN} trained with and without our approach, on the \dataset{Shapes} dataset.\looseness=-1}
  \label{fig:polygons-only-cont}
\end{minipage}
\hfill
\end{figure}

\subsection{Synthetic Datasets}
\label{sec:experiments-shapes}
\vspace{-1ex}
In order to study the properties of our method, we created a synthetic dataset that is easy to analyse, and where a natural coarse-to-fine curriculum might arise.
The questions we investigate are:
(i) how our method performs with varying amounts of training data,
(ii) what the class hierarchy looks like, and
(iii) how the class embedding distance metric compares to other metrics.\looseness=-1 

\paragraph{Dataset Generation.}
We refer to this dataset as \dataset{Shapes}.
The inputs consist of 64$\times$64 images depicting one of 10 geometrical shapes (circles, ellipses, and regular polygons with 3-10 vertices) in one of three colors (magenta, cyan, or grey) against a black background (see Figure~\ref{fig:shapes}).
The dataset contains 50,000 images (5,000 per shape), out of which 10,000 are set aside for testing.
The goal is to predict the shape and its color for each image (i.e., we have 30 classes). 
Our motivation for the design of this dataset is that shapes with similar colors and numbers of vertices look more alike and are thus more likely to confuse the model.
Therefore, we expect our method to help in this setting.\looseness=-1

\paragraph{Results.}
For this dataset, we performed experiments using the \model{CNN} architecture.
Our results, shown in Figure~\ref{fig:polygons-only-cont}, indicate that our method consistently outperforms the baseline. 
Furthermore, we observe that the curriculum method provides the biggest boost over the baseline in the middle regime, when there are not enough samples for the baseline to reach high accuracy, but there is enough to make it a sufficiently good learner that our curriculum learning algorithm can improve upon.
These observations also agree with prior results showing that pre-training is most beneficial in problems where labeled data is scarce.
To further understand where the gains are coming from, we also inspected the generated label hierarchies.
The most common hierarchy generated during our experiments separates all shapes by color on the first level, and by shape similarity on the second level (i.e., circles and ellipses are grouped together, and polygons are also grouped together).
This is intuitive and compatible with what we might have manually constructed.

\input{tables/table-shapes-random.tex}

\paragraph{Distance Metric Evaluation.}
A natural question to ask is whether using the class embedding distance as a distance metric between classes is better than alternative approaches.
For example, what if we force classes that are similar to be separated into different clusters early on (see Appendix~\ref{app:class-sim-measures} for an illustration)?
Could this help the model recognise subtle differences better, by focusing on them early on?
Another natural concern is whether the curriculum itself indeed matters, or the model can simply benefit from any clustering of the labels. This could be because assigning coarse group labels to samples still requires learning feature representations that distinguish these groups (e.g., edges, corners), no matter what the grouping is. 
To answer these questions, we also tested our approach with different distance metrics:
(i) the class confusion matrix\footnote{In practice, we add its transpose to it, since a distance metric needs to be symmetric.} (\dataset{Confusion}),
(ii) a distance matrix defined as 1 minus the confusion matrix (\dataset{ConfusionDist}),
(iii) the class embedding distance used in the rest of our experiments (\dataset{EmbeddingDist}),
(iv) the class embedding similarity (\dataset{EmbeddingSim}) defined as 1 minus embedding distance, and
(v) a random symmetric matrix, whose elements are drawn from a Gaussian distribution with mean 0 and standard deviation 1, which will lead to a random grouping of the classes (\dataset{Random}).
The results are shown in Table~\ref{tab:results-shapes-random}.
We observe that \dataset{Random} hurts performance, as do the metrics that group dissimilar classes early on.
\dataset{EmbeddingDist} and \dataset{ConfusionDist}, which group the most similar classes first, both result in accuracy gains, with the former resulting in the largest gain.
This suggests that using arbitrary hierarchies
is not sufficient; {\em the actual choice of class hierarchy matters}.

\subsection{Real Datasets}
\input{experiments-real.tex}

%% file: tables/table-shapes-random.tex
\begin{wraptable}{h!}{0.5\textwidth}
   \vspace{-2ex}
    \centering
    \caption{
    Results on \dataset{Shapes} with 500 samples per class. 
    We show the accuracy mean and standard error of the baseline, our curriculum approach, and their difference (calculated separately per run and then averaged).}
    \label{tab:results-shapes-random}
    \centering
    \scriptsize
    \lato
    \def\arraystretch{1.1}
    \setlength{\tabcolsep}{1pt}
     \begin{tabularx}{0.5\columnwidth}{|l||Y|Y|Y|}
        \hhline{|-||-|-|-|}
        \multirow{2.5}{*}{\textbf{Distance Metric}} & \multicolumn{3}{c|}{\textbf{Accuracy}} \\
        \hhline{|~||-|-|-|}
        &   \textbf{Baseline} & \textbf{Curriculum} & \textbf{Difference} \\
        \hhline{:=::=:=:=:}
        ConfusionDist      & 39.36$\pm$0.52 & 52.72$\pm$0.98 & \diff{13.49}{1.03} \\
        EmbeddingDist & 39.36$\pm$0.52 & 54.96$\pm$1.37 & \diff{15.70}{1.34} \\
        \hhline{|-||-|-|-|}
        Confusion  & 39.36$\pm$0.52 & 39.59$\pm$1.32 & \diff{0.04}{1.71}  \\
        EmbeddingSim  & 39.36$\pm$0.52 & 34.91$\pm$2.02 & \diff{-4.46}{1.84} \\
        \hhline{|-||-|-|-|}
        Random         & 39.36$\pm$0.52 & 39.00$\pm$1.33 & \diff{-0.32}{1.85} \\
        \hhline{|-||-|-|-|}
    \end{tabularx}
    \vspace{-1ex}
\end{wraptable}

%% file: experiments-real.tex

We performed experiments on the popular datasets \dataset{CIFAR-10}, \dataset{CIFAR-100} \citep{Krizhevsky:2009:cifar-10}, \dataset{Tiny-ImageNet}~\citep{tinyimagenet} and subsets of \dataset{ImageNet}~\citep{Russakovsky:2015:imagenet}.
The \dataset{CIFAR-100} dataset contains labels at two levels of granularity: the original 100 classes, as well as 20 coarse-grained classes.
\dataset{Tiny-ImageNet} is a subset of the \dataset{ImageNet} dataset~\citep{Russakovsky:2015:imagenet} that contains 200 categories and only 500 training examples per category, making it an interesting test case for our method.
Dataset statistics are shown in Table~\ref{tab:dataset-statistics} of Appendix~\ref{app:dataset-stats}.

\begin{figure}[!t]
    \centering
    \includegraphics[width=\textwidth]{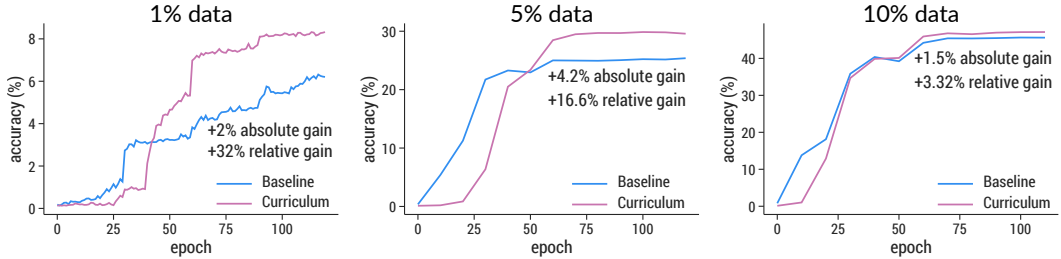}
    \caption{Results for a \model{Resnet18} trained on subsets of \dataset{ImageNet}, on  $1\%$, $5\%$ and $10\%$ of data, respectively.\looseness=-1}
    \label{fig:imagenet}
\end{figure}

\paragraph{Varying the Amount of Training Data.}
Similar to our experiments on synthetic data, we compare our curriculum-based approach to baselines, while varying the number of labeled examples.
\input{tables/results-continuous-real-onlyOurs}

Our results, reported in Table~\ref{tab:results-multiple-sample-size}, show that our approach is able to boost baseline performance for all models and across all datasets.
As expected, the accuracy gains are most significant when we have a large number of labels.
A more controlled setting for observing this effect is to compare the gains on \dataset{CIFAR-100} and \dataset{CIFAR-100 Coarse}, where the input images and the number of samples are similar, but where one dataset contains more fine-grained labels.
Moreover, when comparing results on the same dataset between smaller versus larger models, we observe larger gains for models with lower baseline performance, which is expected given that in these cases there is more room for improvement.
Interestingly, the \model{WideResnet-28-10} performance on \dataset{CIFAR-100} follows the same trend as our observations on the shapes dataset:
we see the largest gains in the middle regime (neither too much nor too little training data). 

We also performed experiments on subsets of \dataset{ImageNet}, on $1\%$, $5\%$ and $10\%$ of the training examples (using all 1000 labels), using \model{Resnet18}.
Due to our limited computational resources, we were not able to perform the same thorough analysis as in Table~\ref{tab:results-multiple-sample-size}.
However, as shown in Figure~\ref{fig:imagenet}, our method is consistently successful at boosting the accuracy of our baselines, with the gains being more prominent in the low data regime.\looseness=-1

\clearpage

\begin{wrapfigure}{r}{0.4\textwidth}
    \centering
    \includegraphics[width=\linewidth]{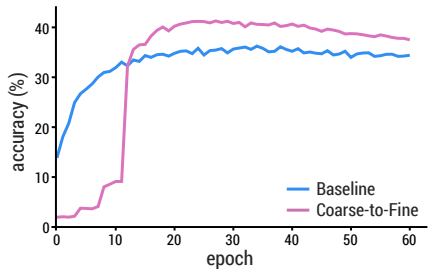}
    \caption{
        Test accuracy per epoch on \dataset{CIFAR-100} using a \model{CNN}. The curriculum had 3 levels, all visited in the first 11 epochs.} 
    \label{fig:acc-per-epoch}
\end{wrapfigure} 

\paragraph{Inspecting the Curriculum.}
To understand how our curriculum has been trained, we show the generated class hierarchy for \dataset{CIFAR-10} in Figure~\ref{fig:overview}, and for \dataset{CIFAR-100} in Appendix~\ref{app:cifar100-hierarchy}.
Most clusters are intuitive, matching our expectation based on semantic similarity (e.g., \class{\{crab, lobster\}}, \class{\{bicycle, motorcycle\}}), but there are a few interesting examples (e.g., \class{\{lamp, cup\}}, \class{\{skyscraper, rocket\}}) that are based on visual similarity as interpreted by the model.
The latter examples stress the importance of automatically generating hierarchies based on what the model itself finds confusing, rather than using hand-crafted ones.
In Figure~\ref{fig:acc-per-epoch}, we show the progression of the accuracy per epoch, with and without curriculum.
Additionally, we include an analysis of how sensitive these results are to the length of the curriculum in Appendix~\ref{app:hyperparam-sensitivity}.


\paragraph{Comparing to Other Approaches.}
We also compare our approach to other related methods:
\begin{enumerate} 
    \item a recent method by \citet{saxena2019data} which is, to the best of our knowledge, the only other existing curriculum approach operating in label space. This method also supports using a curriculum in both the label space and the input space simultaneously. We refer to the two variants of this method as \method{DP-L} (label space) and \method{DP-LI} (label and input space).
    \item a recent hierarchical classification method \citep{nbdt}, termed \method{NBDT}, that uses a label hierarchy generated automatically, similar to our \dataset{EmbeddingSim}.
    \item the popular self-paced learning (\method{SPL}) method of \citet{kumar2010self}, which allows us to compare with an established input space curriculum approach.
    \item a multitask learning approach which trains in parallel all levels of the hierarchy (\method{Multitask}). This is meant to test if training the tasks \textit{sequentially} (as given by the curriculum) matters, or having a contribution from each of them throughout training is enough.
\end{enumerate}

Note that there is no standardized evaluation setting for curriculum learning methods, and thus most of the published approaches use their own custom setup. For a fair comparison, we replicated the setup from \method{NBDT} when training our approach (including the \model{Resnet} and \model{WideResnet} architectures, learning rate and other hyperparameters), and implemented other methods (\method{SPL} and \method{Multitask}) from scratch where code was not available. We also implemented our \model{CNN} architecture within the published code repositories of the methods \method{DP-L} and \method{NBDT}, and ran their training pipeline.
We tried several hyperparameter configurations starting with the ones reported in the original publications (\method{DP-L} and \method{DP-LI} have 3 and 6 curriculum-specific hyperparameters, respectively). For the larger models we report the numbers from their respective publications, where available.

\input{tables/results-baselines}

\footnotetext{This difference is with respect to our baseline accuracy of $64.14\%$, but the difference would be $-0.99\%$ relative to their own baseline,
which we could not reproduce. Similarly, for \dataset{CIFAR-100}, the difference relative to their baseline is 0.78\%.
}

The results are shown in Table~\ref{tab:results-single-label}.
Surprisingly, \method{DP-L} and \method{DP-LI} generally do not perform well for the small \model{CNN}, sometimes even losing performance over the baseline.
They do, however boost the accuracy of the larger model by 0.7\%.
The results for \method{SPL} are inconsistent; it sometimes improve the performance of the baseline and it sometimes does not (note that we have also attempted to tune its pace parameter, denoted as $K$ in the original publication, in order to allow for a fair comparison).
This result is interesting, because \method{SPL} has been shown to work well for other types of problems (though generally involving data of a sequential nature).
One possible explanation for this result is that input space methods such as \method{SPL} are better suited for problems where the difficulty of training examples can be quantified (e.g., some examples involve longer sequences or are noisier than others), and less so for datasets like the ones we consider, where the difficulty is possibly given by the inter-class similarities.
This is possibly why \method{DP-L} and \method{DP-LI} are also very similar in terms of performance.
\method{NBDT} results in some improvements for \model{CNN} and \model{WideResnet}, but also incurs a significant loss in accuracy for \model{Resnet-18}, as also reported in the original paper.
\method{Multitask} generally performs similarly to or slightly better than the baseline, suggesting that there is value in adding a loss function contribution for each level of label granularity.
However, our approach results in the largest improvements, suggesting that learning these tasks {\em sequentially} in a coarse-to-fine order, is important.\looseness=-1

%% file: tables/results-continuous-real-onlyOurs.tex

\begin{wraptable}{r}{0.53\linewidth}
\vspace{1ex}
\caption{
    Results on real datasets, showing the accuracy mean and standard error for the baseline model, computed over 5 runs, as well as the accuracy gain achieved by the our curriculum approach, computed per run and then averaged.
}
\label{tab:results-multiple-sample-size}
\centering
\scriptsize
\lato
\def\arraystretch{1.1}
\setlength{\tabcolsep}{1pt}
\begin{tabularx}{\linewidth}{|c|lr||Y|Y|}
    \hhline{|-|--||--|}
   \multirow{2}{*}{\textbf{Model}} & \multirow{2}{*}{\textbf{Dataset}} & \multirow{2}{*}{\textbf{\#Train}} & \textbf{Baseline Accuracy} & \textbf{Curriculum Acc Gain (\%)} \\
    \hhline{:=:==::==:}
    \multirow{16}{*}{\texttt{CNN}} 
    &CIFAR-10         & 50k                          & 70.92 $\pm$ 0.37 & \diff{0.69}{0.32} \\
    &CIFAR-10         & 20k                          & 64.66 $\pm$ 0.53 & \diff{1.28}{0.60} \\
    &CIFAR-10         & 10k                          & 59.52 $\pm$ 0.35 & \diff{1.24}{0.46} \\
    &CIFAR-10         & {\leavevmode\hphantom{0}}5k  & 53.64 $\pm$ 0.19 & \diff{1.57}{0.39} \\
    \hhline{|~|--||--|}
    &CIFAR-100 Coarse & 50k                          & 49.63 $\pm$ 0.35 & \diff{1.22}{0.38} \\
    &CIFAR-100 Coarse & 20k                          & 42.04 $\pm$ 0.29 & \diff{1.84}{0.51} \\
    &CIFAR-100 Coarse & 10k                          & 36.61 $\pm$ 0.19 & \diff{1.77}{0.56} \\
    &CIFAR-100 Coarse & {\leavevmode\hphantom{0}}5k  & 31.80 $\pm$ 0.28 & \diff{1.38}{0.22} \\
    \hhline{|~|--||--|}
    &CIFAR-100        & 50k                          & 35.87 $\pm$ 0.23 & \diff{3.31}{0.59} \\
    &CIFAR-100        & 20k                          & 27.83 $\pm$ 0.34 & \diff{2.27}{0.37} \\
    &CIFAR-100        & 10k                          & 21.96 $\pm$ 0.49 & \diff{2.67}{0.68} \\
    &CIFAR-100        & {\leavevmode\hphantom{0}}5k  & 17.20 $\pm$ 0.20 & \diff{1.92}{0.24} \\
    \hhline{|~|--||--|}
    &Tiny-ImageNet    & 100k                         & 21.94 $\pm$ 0.19 & \diff{2.73}{0.49} \\
    &Tiny-ImageNet    & {\leavevmode\hphantom{0}}50k & 16.33 $\pm$ 0.32 & \diff{3.06}{0.33} \\
    &Tiny-ImageNet    & {\leavevmode\hphantom{0}}20k & 10.16 $\pm$ 0.22 & \diff{2.02}{0.34} \\
    &Tiny-ImageNet    & {\leavevmode\hphantom{0}}10k & {\leavevmode\hphantom{0}}7.38 $\pm$ 0.11 & \diff{1.14}{0.19} \\
    \hhline{:=:==::==:}
    \multirow{4}{*}{\texttt{Resnet18}}
    & CIFAR-100       & 50k                          & 76.11 $\pm$ 0.20 & \diff{1.08}{0.12} \\
    & CIFAR-100       & 20k                          & 61.24 $\pm$ 0.21 & \diff{2.73}{0.41} \\
    & CIFAR-100       & 10k                          & 46.01 $\pm$ 0.91 & \diff{4.61}{0.40} \\
    & CIFAR-100       & 5k                           & 20.98 $\pm$ 0.35 & \diff{2.32}{0.97} \\
    \hhline{:=:==::==:}
    \multirow{3}{*}{\texttt{Resnet50}}
    & CIFAR-100       & 50k                          & 77.21 $\pm$ 0.40  & \diff{2.20}{0.53} \\
    & CIFAR-100       & 20k                          & 63.31 $\pm$ 0.38 & \diff{0.52}{0.45} \\
    & CIFAR-100       & 10k                          & 51.21 $\pm$ 0.22 & \diff{0.39}{1.01} \\
    \hhline{:=:==::==:}
    \multirow{4}{*}{\texttt{WRN-28-10}} 
    &CIFAR-100      & 50k                            & 80.10 $\pm$ 0.20 & \diff{0.55}{0.13} \\
    &CIFAR-100      & 10k                            & 58.72 $\pm$ 0.38 & \diff{1.29}{0.35} \\
    &CIFAR-100      &  5k                            & 43.77 $\pm$ 0.98 & \diff{2.49}{0.90} \\
    &CIFAR-100      &  1k                            & 14.87 $\pm$ 0.14 & \diff{0.54}{0.56} \\
    \hhline{|-|--||--|}
\end{tabularx}
\vspace{-1ex}
\end{wraptable}

%% file: tables/results-baselines.tex

\begin{table*}[t!]
\centering
\caption{
    Results on real datasets, showing the accuracy mean and standard error for the baseline model, computed over 5 runs, as well as the accuracy gain achieved by the various curriculum approaches, computed per run and then averaged.
    The missing numbers are due to the fact that we were only able to run competing methods using the \model{CNN}, due to limited computational resources.
    For the larger models, we report the numbers published in the respective papers, and do not include standard errors as they were not reported.
    For computational reasons, for the larger models we did not evaluate all competitors (missing results are marked as ``--'').\looseness=-1
}
\label{tab:results-single-label}
\vspace{1ex}
\tiny
\lato
\def\arraystretch{1.3}
\setlength{\tabcolsep}{1.5pt}
\begin{tabularx}{\textwidth}{|c|lc||Y||Y|Y|Y|Y|Y||Y|}
    \hhline{|-|--||-||------|}
    \multirow{2.5}{*}{\textbf{Model}} & \multirow{2.5}{*}{\textbf{Dataset}} & \multirow{2.5}{*}{\textbf{\#Class}} & \textbf{Accuracy (\%)} & \multicolumn{6}{c|}{\textbf{Accuracy Gain (\%)}} \\
    \hhline{|~|~~|:=::=====:t:=:}
    & & & \textbf{Baseline}& \textbf{SPL}      & \textbf{DP-L} & \textbf{DP-LI} &  \textbf{Multitask} &  \textbf{NBDT} & \textbf{C2F (Ours)} \\
    \hhline{:=:==::=::=:=:=:=:=::=:}
    \multirow{4}{*}{\texttt{CNN}}
    & CIFAR-10         & 10  & 70.92{\scriptsize $\pm$}0.37 & \diff{-0.04}{0.19} &  \diff{0.26}{0.20} & \diff{0.53}{0.30} & \diff{0.12}{0.25} & \diff{0.04}{0.38} & \diff{0.69}{0.32} \\
    & CIFAR-100 Coarse & 20  & 49.63{\scriptsize $\pm$}0.35 & \diff{-0.27}{0.09} & \diff{-0.65}{0.34} & \diff{-0.75}{0.47} & \diff{-0.08}{0.24} & --- & \diff{1.22}{0.38} \\
    & CIFAR-100        & 100 & 35.87{\scriptsize $\pm$}0.23 &  \diff{0.94}{0.61} &  \diff{0.14}{0.25} &  \diff{0.26}{0.31} & \diff{0.69}{0.15} & \diff{0.45}{0.45} & \diff{3.31}{0.59} \\
    & Tiny-ImageNet    & 200 & 21.94{\scriptsize $\pm$}0.19 & \diff{-0.97}{0.97} & \diff{-0.05}{0.14} & \diff{-0.08}{0.15} & \diff{0.33}{0.33} & \diff{0.22}{0.28} &  \diff{2.73}{0.49} \\
    \hhline{|-|--||-||-|-|-|-|-||-|}
    \multirow{3}{*}{\texttt{Resnet18}}
    & CIFAR-100 Coarse  & 20 & 84.57{\scriptsize $\pm$}0.14 & \diff{-0.21}{0.82}       & --- & ---               & \diff{0.53}{0.02}                  &  --- & \diff{0.69}{0.11} \\
    & CIFAR-100        & 100 & 76.11{\scriptsize $\pm$}0.20 & \diff{0.51}{0.45} & --- & ---               & \diff{0.01}{0.37} & \dif{-1.19} & \diff{1.08}{0.12} \\
    & Tiny-ImageNet     & 200 & 65.03{\scriptsize $\pm$}0.09 & \diff{-0.19}{0.15}       & --- & ---               &  \diff{-0.76}{0.42} & \dif{-0.80} & \diff{0.12}{0.14} \\
    \hhline{|-|--||-||-|-|-|-|-||-|}
    \multirow{2}{*}{\texttt{Resnet50}}
    & CIFAR-100 Coarse &  20 & 84.68{\scriptsize $\pm$}0.47 & \diff{-1.21}{0.56}                 & --- & ---               & \diff{-0.97}{0.51}                  & --- & \diff{0.49}{0.41} \\
    & CIFAR-100        & 100 & 77.21{\scriptsize $\pm$}0.40              & \diff{-1.68}{0.55}       & --- & ---               &  \diff{0.43}{0.89} & --- & \diff{2.20}{0.53} \\
    \hhline{|-|--||-||-|-|-|-|-||-|}
    \multirow{2}{*}{\texttt{WRN-28-10}}
    & CIFAR-100         & 100 & 80.10{\scriptsize $\pm$}0.20 & \diff{0.50}{0.41}                 & --- & \diff{0.70}{0.33} & \diff{0.21}{0.12} & \dif{2.77}  & \diff{0.55}{0.13} \\
    & Tiny-ImageNet    & 200 & 64.14{\scriptsize $\pm$}0.15            & ---                & --- & ---               &      ---       & \dif{2.52}\footnotemark{} & \diff{1.01}{0.22} \\
    \hhline{|-|--||-||-|-|-|-|-||-|}
\end{tabularx}
\vspace{-2ex}
\end{table*}

%% file: related-work.tex
\section{Related Work}
\label{sec:related-work}

Most prior work in curriculum learning focuses on the notion of example difficulty, rather than task difficulty~\citep[e.g.,][]{Wang:2018:mancs,Zhou:2018:minimax, guo2018curriculumnet, Jiang:2015:self-paced-curriculum,Jiang:2018:mentor-net,Bengio:2009:curriculum}.
The difficulty of the examples is estimated based on problem-specific rules \citep[e.g., sentence length in natural language processing;][]{Platanios:2019:competence, Bengio:2009:curriculum} or based on the progress of the learner \citep[e.g., the loss on each sample in SPL;][]{kumar2010self, Jiang:2015:self-paced-curriculum}.
Using the sample difficulties, these methods then decide when a sample should be shown to the model, starting with the easy ones first.
Perhaps most related to our work is the work of \citet{saxena2019data}, where the logits predicted for each training example are scaled by a corresponding class weight, which is learned together with the model parameters. 
While this can be seen as a form of curriculum in output space, the core idea is different.\looseness=-1

There also exist a few methods that consider curricula in task space in the context of multitask learning~\citep[e.g.,][]{pentina:2015:curriculum-multitask,guo2018dynamic,sarafianos2018curriculum}.
However, multitask learning is different than our setting because:
(i) the tasks are provided, rather than being automatically generated, and
(ii) the goal is typically to perform well on all of the tasks and not just one of them.
A line of work related to curriculum in {\em output} space and which can be viewed as a form of curriculum in {\em task} space is in the area of reinforcement learning, where agents are trained to achieve incrementally more difficult goals~\citep[e.g.,][]{sukhbaatar2017intrinsic, Florensa2017ReverseCG, Svetlik2017AutomaticCG, Narvekar2017AutonomousTS}.
While these approaches do alter the target task, they cannot be directly translated to the supervised classification regime.\looseness=-1

The idea of solving tasks in a coarse-to-fine order has previously been explored in computer vision and signal processing \citep[e.g., for object detection and recognition][]{fleuret2001coarse,amit2004coarse,moreels2005probabilistic,ren2018generalized}, head pose estimation \citep{wang2019deep}, or more general computer vision tasks \citep{lu2011coarse,mazic2015two,sahbi2002coarse,wu2019liteeval,zambanini2012coarse}.
Similarly, coarse-to-fine ideas have also been used for various tasks in natural language processing \citep[e.g.,][]{lee-etal-2018-higher,dong-lapata-2018-coarse,yao2019coarse}.
Our approach is different in that it is widely applicable; it does not depend on the problem space at hand, but can rather be applied as-is to any classification problem.
In a different line of work, \citet{srivastava2013discriminative} proposed a probabilistic learning method using a class hierarchy.
A tree-based prior over the last layer of a neural network is used to encourage classes closer in the hierarchy to have similar parameters.
Moreover, \citet{bilal2017convolutional} obtain coarse clusters of classes from the confusion matrix and introduce extra branches to a neural network architecture during training that classify the coarse classes simultaneously during training (similar to our \method{Multitask} baseline, but the branches are introduced at custom positions in the model).\looseness=-1

Finally, some of the ideas in our paper bear some resemblance to hierarchical classification \citep[HC; e.g.,][]{bennett2009refined,ramaswamy2015convex,ramirez2016hierarchical,xu2019hierarchical}.
As discussed in Section~\ref{sec:method-analysis}, there is a fundamental difference between our approach, which is purely a training algorithm, and the various HC methods (that also leverage the hierarchy during inference, such as NBDT \citep{nbdt}).
For example, some HC methods propose special architectures that can make predictions at all levels of the hierarchy and combine them \citep[e.g.,][]{wehrmann2018hierarchical}, or use the predictions for coarse labels as inputs to a classifier for finer labels \citep[e.g.,][]{bennett2009refined}.
We provide a more detailed comparison with different HC approaches in Appendix~\ref{app:hierarchical-classification}.
However, the important take-away is that our method is:
(i) intended to be general purpose, meaning that it can be used to train any baseline model without requiring a special architecture, and
(ii) does not affect the model during inference, meaning that it does not require any extra memory or computation. 

%% file: conclusion.tex
\section{Conclusion}

We proposed a curriculum learning algorithm for classification that:
(i) breaks down complex classification tasks into sequences of simpler tasks, and
(ii) goes through these tasks in order of increasing difficulty, training classifiers and transferring acquired knowledge between the learned classifiers.
We showed that our approach achieves significant performance gains on both synthetic and real data, using multiple neural network architectures.
Finally, our approach is purely a training strategy, and does not incur any additional memory or computational costs during inference.

A note relating to ethical concerns: our approach is a training strategy that can be used to train classifiers in any application area. It does not depend on, nor did we use in our experiments, any data that may cause ethical concerns. However, due to its generality, the users of our method may apply it to any classification task of their choice, and it thus bears the same risks of being misused as any other training algorithm.

%% file: appendix.tex
\appendix
\newpage
\section*{Appendix}
\addcontentsline{toc}{section}{Appendices}
\renewcommand{\thesubsection}{\Alph{subsection}}

\subsection{Confusion Matrix vs Embedding Similarity}
\label{sec:confusion-degenerate-case}

In Section~\ref{sec:auxiliary-tasks}, we considered two different measures of class similarity: one based on the confusion matrix and one based on the class embedding distance.
We experimented with both and observed a few disadvantages to using the confusion matrix.
This made us opt for a measure that is based on the class embedding distance.
In what follows, we discuss these disadvantages.
We recommend reading Section~\ref{sec:auxiliary-tasks} before proceeding with this section, as some of the issues discussed here are related to the way we intend to use the class similarity measure.

First, we need to define how the confusion matrix is estimated from the data.
We define the confusion matrix as $\mathbf{C} \in [0, 1]^{K \times K}$, where $K$ is the number of classes and $\mathbf{C}_{ij}$ is the probability that the model predicts class $j$ when it should have predicted class $i$, and  $\smash{\sum_{j=1}^K \mathbf{C}_{ij}=1}$.
Given an existing model, this matrix can be approximated using the sample estimate of each probability on a validation dataset.
Using $\mathbf{C}$ as similarity between the classes in the hierarchical clustering algorithm, we encountered the following issues:

\begin{itemize}[label=--, topsep=0pt]
    \item If the training set is imbalanced, and one class dominates in the number of training examples, it is possible that the classifier often confuses all other classes for the dominating class, instead of mistaking them for more semantically similar classes.
    This is because making such a mistake during training is likely to incur a lower loss. Thus, using the confusion matrix as similarity measure, the most similar class to all other classes could be the dominating class, regardless of its semantics. As a consequence, affinity clustering~\citep{Bateni:2017:affinity} will connect all classes to the dominating class in the first level of the hierarchy, resulting in a degenerate case with no auxiliary functions. 

    \item The confusion matrix is not a proper distance metric, as is not symmetric and does not necessarily satisfy the triangle inequality.
    Although this did not necessary pose a problem for our implementation of the affinity clustering algorithm, being a proper distance metric is important for other hierarchical clustering algorithms, and could cause issues if the users of our algorithm chose to use a different clustering method.
    In our experiments, we made it symmetric by adding its transpose to itself (i.e. $\mathbf{C} + \mathbf{C}^\top$ ).

    \item If the classifier does not confuse two classes $i$ and $j$ at all in the validation set, their confusion count will be $0$, and they will thus be considered highly dissimilar.
    Using an example from \dataset{CIFAR-100}, the classes \texttt{willow\_tree}, \texttt{oak\_tree}, \texttt{palm\_tree} and \texttt{pine\_tree} are similar semantically and so, intuitively we would expect them to be grouped in the same cluster.
    However, suppose that in our validation set \texttt{willow\_tree} and \texttt{palm\_tree} are always confused only with each other, and the same happens for \texttt{oak\_tree} and \texttt{pine\_tree}.
    Then \texttt{willow\_tree} and \texttt{palm\_tree} classes will be grouped together early on, and \texttt{oak\_tree} and \texttt{pine\_tree} will also be grouped together early.
    After this grouping, the confusion between the two new clusters will be $0$ and so they will only be merged at the top of the hierarchy.
    The class embedding similarity, being the cosine distance between two high dimensional vectors, does not suffer from the same issue.
\end{itemize}






\subsection{A Staged Coarse-to-Fine Approach}
\label{app:staged}

As mentioned in Section~\ref{sec:knowledge-transfer}, there is also another version of our algorithm, where we can train a different classifier $f_{\theta_{\ell}}$ at each level $\ell$ of the hierarchy,  which projects directly to the coarse classes at level $\ell$ . 
We presented this staged approach in our workshop paper \citep{Stretcu:2020:coarse-to-fine}, and we summarize it here and compare it with the continuous version of our approach.

The main difficulty in this case is that $f_{\theta_{\ell+1}}$ and $f_{\theta_\ell}$ make predictions for different number of classes, and thus the number of parameters in $\theta_{\ell+1}$ does not directly match that of $\theta_{\ell}$.
Let us assume that for any level $\ell$,
\begin{equation}
f_{\theta} = \underbrace{f^H_{\theta^H}}_{\mathclap{\code{predictor}}} \circ \underbrace{f^{H-1}_{\theta^{H-1}} \circ \cdots \circ f^1_{\theta^1}}_{\code{encoder}},
\end{equation}

\noindent where $\circ$ denotes function composition, $H$ is the number of layers in the network, and $\theta =\{ \theta^1, \hdots, \theta^H \}$.
This is a simple decomposition that applies to most deep learning models that are used in practice.
Intuitively, the encoder converts its input to a latent representation (i.e., embedding), that is then processed by the predictor to produce a probability distribution over classes.
Let us further denote the parameters of the predictor by $\smash{\theta^{\code{pred}}}$ and those of the encoder by $\smash{\theta^{\code{enc}}}$ (we have that $\theta = \theta^{\code{pred}} \cup \theta^{\code{enc}}$).
We suggest decomposing $f_{\theta}$ such that most of the model parameters are part of the encoder.
In our experiments, the predictor is simply the output layer of a neural network, whose output dimensionality changes at every hierarchy level, depending on the number of clusters in that level.
When training $f_{\theta_{\ell+1}}$, we initialize its encoder parameters as $\smash{\theta_{\ell+1}^{\code{enc}} = \theta_{\ell}^{\code{enc}}}$, and its predictor parameters $\smash{\theta_{\ell+1}^{\code{pred}}}$ randomly.
Thus, knowledge transfer in this case happens through the initialization of the encoder parameters, which often include most of the model parameters.
The main intuition behind this decision is that lower level processing (e.g., converting pixels to edges and potentially to abstract semantic features) is a step that is necessary for most levels of granularity.
However, the predictor parameters are specific to the each task (i.e., the predictor decides how the higher-level features are assembled together to solve each task).
$\theta_1$ is initialized randomly in our experiments. 
Putting the pieces together, first we generate a class hierarchy as described in Algorithm~\ref{alg:generating-class-hierarchy}, and then we train a classifier at each level of the hierarchy, transferring knowledge via the model parameters as described above.
These steps are detailed in Algorithm~\ref{alg:coarse-to-fine-curriculum-staged}.
We refer to this approach as the {\em staged} variant of our curriculum learning algorithm, and we refer to the approach discussed in the main paper as the {\em continuous} variant.

\input{algorithms/algo-curriculum-staged}


The two proposed training algorithms, staged and continuous, come with advantages and disadvantages. Here we discuss the trade-offs in terms of hyperparameters and computational complexity.

\paragraph{Hyperparameters.}
An advantage of the staged approach is that it introduces no extra hyperparameters that we need to tune.
The number of levels in the hierarchy is automatically determined by the output of the hierarchical clustering algorithm and we train until convergence for each level.
The continuous approach introduces a hyperparameter, which is total number of epochs to be spent on the curriculum, $T$, and which is then split equally among the levels of the hierarchy.

\begin{figure}[t]
    \centering
    \includegraphics[height=5cm]{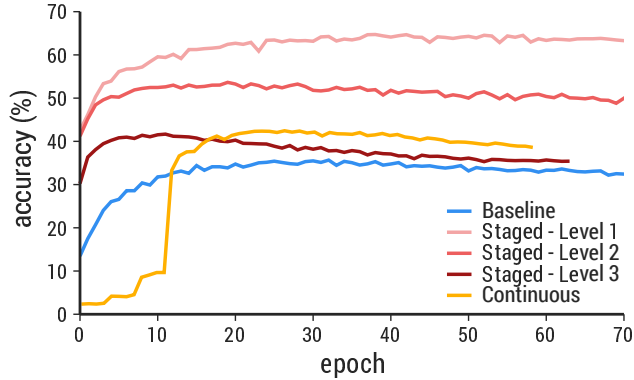}
    \caption{
        Accuracy per epoch for the baseline and our algorithm, on the \dataset{CIFAR-100} dataset.}
    \label{fig:acc-per-epoch-with-staged}
\end{figure}

\paragraph{Computational Complexity.}
Let $\mathcal{C}$ be the computational complexity required to train the baseline model to convergence.
Since affinity clustering guarantees that our class hierarchy will have at most $\log K$ levels, where $K$ is the original number of classes, the computational complexity of the staged approach will be at most $\mathcal{C}\log K$.
However, in practice we observed that the cost is significantly less for two reasons:
(i) training the coarse-grained classifiers converges much faster than the fine-grained baseline, and
(ii) after $f_{\theta_1}$, all other levels are already pre-trained by being initialized with the parameters from the previous level, and thus require very few training iterations.
Evidence of this behavior can be seen in Figure~\ref{fig:acc-per-epoch-with-staged}.
As discussed in Section~\ref{sec:method-analysis}, our continuous curriculum has roughly the same computational complexity as the baseline.
Using our heuristic for setting the number of epochs $T$, the continuous curriculum typically requires about as many training iterations as the baseline model.

\begin{figure}[t]
  \centering
    \captionsetup{width=.7\linewidth}
    \includegraphics[height=5cm]{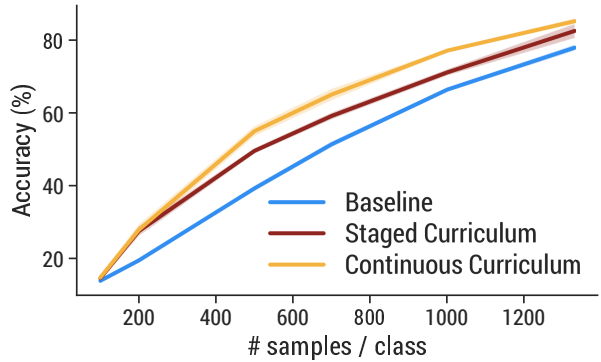}
    \caption{
        Accuracy mean and standard error for the baseline and the curriculum model, averaged over 5 runs, on \dataset{Shapes}.}
    \label{fig:polygons-with-staged}
\end{figure}

\paragraph{Experiments using Staged Coarse-to-Fine Curriculum.}
We show results for the staged curriculum, on similar settings to the experiments reported in the main paper, using the \model{CNN} model.
Note that, for all these experiments, we do not use any image augmentation techniques or specialized learning rate schedules, since we wanted to understand the effect of our methods without extra help from such techniques.
The results are reported in Table~\ref{tab:results-single-label-staged} and Figures \ref{fig:acc-per-epoch-with-staged} and \ref{fig:polygons-with-staged}.
Table~\ref{tab:results-single-label-staged} shows that the staged approach also provides a significant boost over the baseline method, occasionally even better than the continuous approach (albeit at a larger computational cost).
Figure~\ref{fig:acc-per-epoch-with-staged} shows the training curve of the staged curriculum at each level of the hierarchy, as well as a comparison with the baseline and the continuous method shown in Figure~\ref{fig:acc-per-epoch}.
Importantly, even for \dataset{CIFAR-100} where we have 100 labels, the hierarchy only contains two auxiliary levels with $6$ and $27$ clusters, respectively.
This means that the staged approach can achieve accuracy improvements in the order of 3-4\% with at most 3 times the computational cost.
In practice, the actual cost is much less than that, because each hierarchy level now needs much fewer iterations to converge than the baseline, as shown in Figure~\ref{fig:acc-per-epoch-with-staged}.
This figure also confirms our intuition that the auxiliary tasks obtained with our approach are indeed sorted in order of difficulty, in the sense that the accuracy of the model at solving levels 3, 2, and 1 is monotonically increasing.
Additionally, we directly notice the benefits of pre-training by looking the accuracy of the staged models after one epoch.

\input{tables/results-staged-real}

\subsection{Experimental Details}
\label{app:exp-details}

\paragraph{Architecture Details.}
The Convolutional Neural Network (\model{CNN}) used in our experiments consists of these layers:
\begin{enumerate}[itemsep=0.5ex,leftmargin=2em]
    \item \uline{Convolution:} 2D convolution using a $3 \times 3$ filter with 32 channels, followed by ReLU activation.
    \item \uline{Pooling:} Max pooling using a $2 \times 2$ window.
    \item \uline{Convolution:} 2D convolution using a $3 \times 3$ filter with 64 channels, followed by ReLU activation.
    \item \uline{Pooling:} Max pooling using a $2 \times 2$ window.
    \item \uline{Convolution:} 2D convolution using a $3 \times 3$ filter with 64 channels, followed by ReLU activation.
    \item \uline{Projection:} Fully connected layer performing a linear projection to the output space dimensionality (i.e., number of classes), returning logits.
\end{enumerate}
The \model{WideResnet-28-10} and \model{Resnet18} architectures were implemented after the code released by \citet{nbdt}, and are similar to the original publications \citep{Zagoruyko2016WideRN, he2016deep}.

\paragraph{Training.}
We implemented our method using the TensorFlow framework~\citep{abadi2016tensorflow}.
All models were trained by minimizing the softmax cross-entropy loss function.

All the \model{CNN} experiments used the Adam optimizer~\citep{kingma2014adam} with a learning rate of 0.001 and a batch size of 512 samples.
We also employed early stopping by terminating training when validation accuracy did not improve within the last $50$ epochs (we made sure that this number is large enough by visually inspecting the validation curve).
We also made sure that the baseline is allowed to perform at least as many epochs as the curriculum.
We report test accuracy statistics for the iteration that corresponds to the best validation set performance.
The validation dataset is obtained by setting aside 20\% of the training examples, chosen uniformly at random.

For the \model{Resnet} and \model{WideResnet} experiments we replicated the setting of \citet{nbdt}, available at {\small \url{https://github.com/alvinwan/neural-backed-decision-trees}}, in order to be able to directly compare with their results.
Concretely, we used a Stochastic Gradient Descent (SGD) optimizer with momentum 0.9, batch size 128, and weight decay value of 5e-4, and trained the models for 200 epochs.  
The learning rate schedule starts with a learning rate of 0.1 and is divided by 10 twice: $\frac{3}{7}$ and $\frac{5}{7}$ of the way
through training.
Because of this step-wise learning rate, our heuristic for choosing the curriculum length (when the baseline has reached $90\%$ of its peak accuracy) no longer applies, so in this case the curriculum length was chosen based on validation set performance among the choices \{5, 10, 20, 30, 40, 50\}. First we randomly split the training set in a 90\% train and 10\% validation and trained the models with each of these curriculum lengths. We then chose the best performing curriculum length on the validation set, and retrained on the full training set using this length.
Importantly, in all our experiments we only allowed both the baseline and the corresponding curriculum models to train for {\em exactly the same number of epochs}.

To match the implementation of \citet{nbdt}, for the large models we used the same data augmentation techniques they did, while for the \model{CNN} experiments we opted out of data augmentation to see just how much the curriculum impacts this simple model.

Finally, all our experiments were performed using a single Nvidia Titan X GPU. 

\subsection{Class Similarity Metrics}
\label{app:class-sim-measures}

\begin{figure}[h!]
    \centering
    \includegraphics[width=\linewidth]{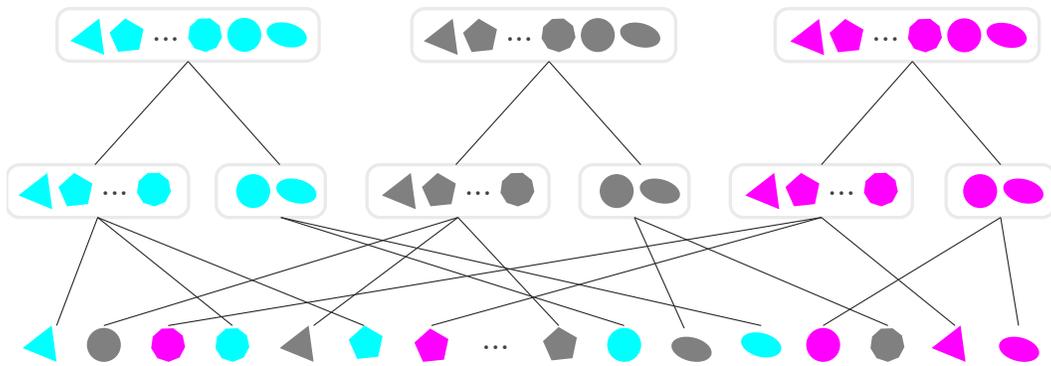}
    \vspace{1ex}
    \caption{Hierarchy generated using \dataset{EmbeddingDist} as class distance measure. The hierarchy is created bottom-up, starting by first connecting the shapes that have the lowest embedding \textit{distance} (i.e. those that are most similar) at the bottom of the hierarchy.}
    \label{fig:hierarchy-easy-to-hard}
\end{figure}

\begin{figure}[h!]
    \centering
    \includegraphics[width=\linewidth]{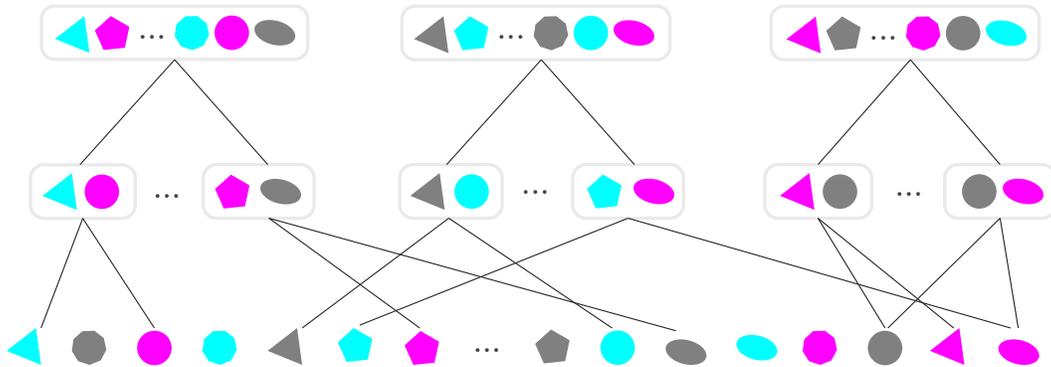}
    \vspace{1ex}
    \caption{Hierarchy generated using \dataset{EmbeddingSim} as class distance measure. The hierarchy is created bottom-up, starting by first connecting the shapes that have the lowest embedding \textit{similarity} (i.e. those that are most dissimilar) at the bottom of the hierarchy.}
    \label{fig:hierarchy-hard-to-easy}
\end{figure}

In Section \ref{sec:experiments-shapes} we discuss different ways to obtain a label hierarchy, by changing the class distance measure provided as input the the agglomerative clustering algorithm. By using either \dataset{ConfusionDist} or \dataset{EmbeddingDist} as distances between classes we obtain a hierarchy similar to Figure \ref{fig:hierarchy-easy-to-hard}, where the most similar classes (e.g. gray circle and gray ellipse) are separated in the very last level of the hierarchy (i.e. later on during training, according to our coarse-to-fine curriculum). On the contrary, reverting these distances, and using \dataset{Confusion} or \dataset{EmbeddingSim} as inputs to the clustering algorithm, leads to a hierarchy as shown in Figure \ref{fig:hierarchy-hard-to-easy}, where the most similar classes are separated at the first level of the hierarchy.

\subsection{Dataset information}
\label{app:dataset-stats}


Table~\ref{tab:dataset-statistics} shows some statistics for all datasets used in our experiments.
The data and specific train/test splits for \dataset{CIFAR-10}, \dataset{CIFAR-100}, and \dataset{CIFAR-100 Coarse} were obtained from the \texttt{tensorflow\_datasets} package \footnote{\url{https://www.tensorflow.org/datasets/catalog/}} from Tensorflow.
For \dataset{Tiny-ImageNet}, we used the provided train/validation/test splits as \citet{tinyimagenet}.

\begin{table}[!h]
    \vspace{-2ex}
    \caption{Statistics for the multi-class classification datasets used in our experiments.}
    \label{tab:dataset-statistics}
    \centering
    \scriptsize
    \lato
    \def\arraystretch{1.2}
    \setlength{\tabcolsep}{3pt}
    \begin{tabularx}{\linewidth}{|l||Y|Y|Y|}
        \hhline{|-||-|-|-|}
        \textbf{Dataset} & \textbf{\# classes} & \textbf{\# train} &  \textbf{\# test} \\
        \hhline{:=::=:=:=:}
        Shapes           & {\leavevmode\hphantom{0}}30  & {\leavevmode\hphantom{0}}40,000  & 10,000  \\
        CIFAR-10         & {\leavevmode\hphantom{0}}10  & {\leavevmode\hphantom{0}}50,000  & 10,000 \\
        CIFAR-100 Coarse & {\leavevmode\hphantom{0}}20  & {\leavevmode\hphantom{0}}50,000  & 10,000 \\
        CIFAR-100        &                          100 & {\leavevmode\hphantom{0}}50,000  & 10,000 \\
        Tiny-ImageNet    &                          200 &                          100,000 & 10,000 \\
        \hhline{|-||-|-|-|}
    \end{tabularx}
\end{table}



\subsection{Generated label hierarchy for CIFAR-100}
\label{app:cifar100-hierarchy}
\begin{itemize}[leftmargin=2em]
    \item \underline{Level 1:}
        \begin{enumerate}[label={Cluster \arabic* :},leftmargin=6em,noitemsep]
            \item apple, pear, sweet\_pepper, orange, aquarium\_fish, sunflower, rose,
          orchid, tulip, poppy, crab, lobster
            \item  baby, woman, girl, hamster, boy, man, fox, lion, snail, camel
            \item flatfish, ray, shark, turtle, dolphin, whale, bear, chimpanzee, skunk,
          cattle, dinosaur, elephant, seal, otter
        \item  crocodile, lizard, shrew, beaver, porcupine, mushroom, kangaroo, tiger,
          leopard, trout, possum, wolf, mouse, raccoon, squirrel, rabbit
        \item lamp, cup, worm, chair, bed, table, keyboard, couch, snake, bicycle,
          motorcycle, can, telephone, television, bottle, wardrobe, bowl, plate, clock
        \item caterpillar, bee, butterfly, cockroach, spider, beetle
       \item  willow\_tree, forest, oak\_tree, palm\_tree, pine\_tree, maple\_tree, skyscraper,
          rocket, tractor, train, tank, castle, bridge, house, streetcar, pickup\_truck,
          bus, lawn\_mower, mountain, cloud, road, sea, plain
        \end{enumerate}
  \item \underline{Level 2:}
  \begin{enumerate}[label={Cluster \arabic* :},leftmargin=6em,noitemsep]
        \item apple, pear, sweet\_pepper, orange
        \item aquarium\_fish, sunflower, rose, orchid, tulip, poppy
        \item bowl, plate, clock
        \item castle, bridge, house
        \item streetcar, pickup\_truck, bus, lawn\_mower
        \item fox, lion, snail, camel
        \item skunk, cattle, dinosaur, elephant
        \item mountain, cloud, road, sea, plain
        \item crab, lobster
        \item crocodile, lizard
        \item lamp, cup
        \item flatfish, ray, shark, turtle, dolphin, whale
        \item baby, woman, girl, hamster, boy, man
        \item willow\_tree, forest, oak\_tree, palm\_tree, pine\_tree, maple\_tree
        \item mushroom, kangaroo, tiger, leopard, trout
        \item possum, wolf, mouse, raccoon
        \item seal, otter
        \item squirrel, rabbit
        \item skyscraper, rocket
        \item tractor, train, tank
        \item bear, chimpanzee
        \item shrew, beaver, porcupine
        \item worm, chair, bed, table, keyboard, couch, snake
        \item caterpillar, bee, butterfly
        \item cockroach, spider, beetle
        \item bicycle, motorcycle
        \item can, telephone, television, bottle, wardrobe
    \end{enumerate}
    \item \underline{Level 3}: Each class has its own cluster. 
\end{itemize}

\begin{figure*}[t]
    \centering
    \includegraphics[width=0.8\linewidth]{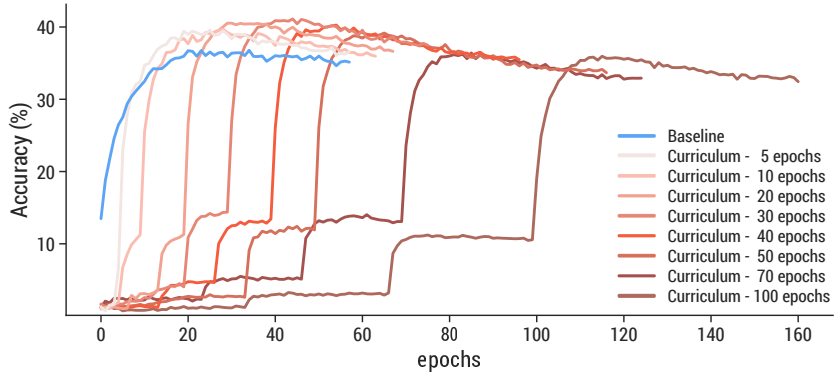}
    \caption{Accuracy of our coarse-to-fine curriculum strategy with different curriculum lengths on \dataset{CIFAR-100}.}
    \label{fig:sensitivity}
\end{figure*}

\subsection{Hyperparameter Sensitivity}
\label{app:hyperparam-sensitivity}
We evaluate the sensitivity of our curriculum approach to the length of the curriculum. In Figure~\ref{fig:sensitivity}, we show the accuracy of the \model{CNN} on the \dataset{CIFAR-100} dataset, using various curriculum lengths.
For a fixed curriculum length (in number of epochs), we divide it equally among the hierarchy levels. Once the final hierarchy level is reached, we train until the validation accuracy has not improved for the last 50 epochs. As can be seen in Figure~\ref{fig:sensitivity}, this threshold is enough for both the baseline model and the curriculum models to reach their best point.
The results suggest that there is an optimal middle curriculum length: too few epochs and the coarse hierarchy levels do not have enough time to train and provide good initial points for the next level; too many epochs and the auxiliary hierarchy levels start overfitting, and their performance starts dropping before we move on to the next level.
However, our curriculum method is robust overall, because despite the large range of curriculum lengths we used across all experiment runs, the performance of the curriculum trained model consistently outperformed or at least matched the baseline model performance.

\subsection{Other Related Work}
\label{app:hierarchical-classification}

In the main paper, we discussed how our approach may be related to hierarchical classification methods.
Therefore, we expand here on the main directions in hierarchical classification and compare these with our method.
A popular survey on hierarchical classification by \citet{silla2011survey} organizes the hierarchical classification literature in three main types of approaches:
\begin{itemize}[label=--,leftmargin=1.5em, topsep=0pt]
    \item \textit{flat classification approaches}, which ignore the class hierarchy and consider only the fine-grained leaf-node classes. This is equivalent to any standard classification problem.
    \item \textit{local classification approaches}, which typically go top-down through the label hierarchy and train multiple classifiers along the way that take into account only local information \citep[e.g.,][]{bennett2009refined, ramaswamy2015convex, ramirez2016hierarchical}. These approaches use various ways of incorporating local information, such as having a classifier per node~\citep{jin2008multi, valentini2009weighted}, per parent node~\citep{gauch2009training}, or per hierarchy level~\citep{clare2003predicting}. Knowledge can be passed down through the hierarchy in various ways, for example using the predictions of the parent node as input to the current classifier \citep{holden2009hierarchical, bennett2009refined}.
    A more recent approach~\citep{xu2019hierarchical} is based on the idea that two labels with a common ancestor in the class hierarchy are correlated, and explicitly models this correlation in the label distribution.
    Approaches in this category are typically computationally more expensive, and tend to be sensitive to error propagation along the hierarchy levels.
   \item \textit{global classification approaches}, which train a single classification function that takes into account the entire class hierarchy at once \citep{cai2004hierarchical,cerri2012genetic,  wang2009large, Xiao2011HierarchicalCV}.
   For example, \citet{cai2004hierarchical} do so by designing a generalization of Support Vector Machines (SVM) using discriminant functions that decompose into contributions from different levels of the hierarchy. \citet{Xiao2011HierarchicalCV} train a hierarchical SVM which consists of a classifier at each node, but information about the whole hierarchy is encoded in a regularization term that encourages the normal vector of the classifying hyperplane at each node to be orthogonal to those of its ancestors.
   The more recent work of \citet{wehrmann2018hierarchical} proposes a new neural network architecture for class hierarchies, which can make predictions at different levels of the hierarchy and is trained by combining multiple losses: a local loss, a global loss, and a loss that penalizes predictions that violate the hierarchy. This approach is not merely a training mechanism for training arbitrary models, but is tied together to the proposed architecture.
\end{itemize}
Our approach is similar to local node classification approaches in that we also train a classifier at each level.
It is also in some sense similar to global classification approaches, because at the end of training, the model at the final level of the hierarchy is able to preserve knowledge about the rest of the hierarchy through the parameters that have been propagated through the levels.
However, to the best of our knowledge, none of these hierarchical classification approaches pass information across levels only through the model parameters. Typically this information is encoded directly in the model itself, which often means that the hierarchy is also used during inference, not just during training.

%% file: algorithms/algo-curriculum-staged.tex
{
\centering
	\begin{algorithm2e}[t]
	\caption{Coarse-To-Fine Curriculum: A Staged Approach}
	\label{alg:coarse-to-fine-curriculum-staged}
	\footnotesize
	\tcp{\code{\scriptsize This is an overview of the proposed staged curriculum algorithm.}}
	\SetKwInOut{Input}{Inputs}
	\Input{Number of classes $K$.\\
	Training data $\{x_i, y_i\}_{i=1}^N$.\\
	Trainable baseline model $f_{\theta}$.}
	Train $f_{\theta}$ on the provided training data $\{x_i, y_i\}_{i=1}^N$.\\
	$\code{clustersPerLevel} \leftarrow \code{\textcolor{myBlue}{GenerateClassHierarchy}(}$ $K, \{x_i, y_i\}_{i=1}^N, f_{\theta}\code{)}$\\
	$\code{M} \leftarrow  \code{clustersPerLevel.length} $\\ 
	\tcp{Train the model at each level of the hierarchy.}
	$\code{originalLabels} \leftarrow \code{[1,...,K]}$\\
	\For{$l \leftarrow 0,\hdots,\code{M - 1}$}{
	    $\code{clusters} \leftarrow \code{clustersPerLevel[}l + 1\code{]}$\\
	    $\code{newLabels} \leftarrow \code{\textcolor{myBlue}{TransformLabels}(}$ $\{y_i\}_{i=1}^N\code{, clusters)}$\\
	    \uIf{$l = 0$}{$\theta_{l+1}^{\code{encoder}} \leftarrow \code{random()}$.}
	    \Else{$\theta_{l+1}^{\code{encoder}} \leftarrow \theta_{l}^{\code{encoder}}$}
	    $\theta_{l+1}^{\code{predictor}} \leftarrow \code{random()}$.\\
	    Train $f_{\theta_{l+1}}$ using $\code{newLabels}$ as the target labels.
	}
	\KwOut{$f_{\theta_{\code{[\code{M}]}}}$.}
	\afterpage{\global\setlength{\textfloatsep}{\oldtextfloatsep}}
	\end{algorithm2e}
\par
}

%% file: tables/results-staged-real.tex
\begin{table*}[t!]
\centering
\caption{
    Results on real datasets using the \model{CNN} architecture, showing the accuracy mean and standard error for the baseline model, computed over 5 runs, as well as the accuracy gain achieved by the two versions of our coarse-to-fine curriculum (staged and continuous), computed per run and then averaged. Note that here we do not use any image augmentation techniques or specialized learning rate schedules.}
\scriptsize
\lato
\def\arraystretch{1.1}
\setlength{\tabcolsep}{3pt}
\begin{tabularx}{\textwidth}{|lcc||Y||Y|Y|}
    \hhline{|---||-||--|}
    \multirow{2.5}{*}{\textbf{Dataset}} & \multirow{2.5}{*}{\textbf{\#Class}} & \multirow{2.5}{*}{\textbf{\#Samples}} & \textbf{Accuracy} & \multicolumn{2}{c|}{\textbf{Accuracy Gain}} \\
    \hhline{|~~~|:=::==:}
             &    &        & \textbf{Baseline}& \textbf{Coarse-to-Fine-Staged} & \textbf{Coarse-to-Fine-Continuous} \\
    \hhline{:===::=::=:=:}
    CIFAR-10 & 10 & 50,000 & 70.92 $\pm$ 0.37 & \diff{0.92}{0.32} & \diff{0.69}{0.32} \\
    CIFAR-10 & 10 & 20,000 & 64.66 $\pm$ 0.53 & \diff{1.86}{0.23} & \diff{1.28}{0.60} \\
    CIFAR-10 & 10 & 10,000 & 59.52 $\pm$ 0.35 & \diff{1.01}{0.13} & \diff{1.24}{0.46} \\
    CIFAR-10 & 10 & {\leavevmode\hphantom{0}}5,000 & 53.64 $\pm$ 0.19 & \diff{2.22}{0.42} & \diff{1.57}{0.39} \\
    \hhline{|---||-||-|-|}
    CIFAR-100 Coarse & 20 & 50,000 & 49.63 $\pm$ 0.35 & \diff{0.91}{0.37} & \diff{1.22}{0.38} \\
    CIFAR-100 Coarse & 20 & 20,000 & 42.04 $\pm$ 0.29 & \diff{1.03}{0.22} & \diff{1.84}{0.51} \\
    CIFAR-100 Coarse & 20 & 10,000 & 36.61 $\pm$ 0.19 & \diff{1.38}{0.53} & \diff{1.77}{0.56} \\
    CIFAR-100 Coarse & 20 & {\leavevmode\hphantom{0}}5,000 & 31.80 $\pm$ 0.28 & \diff{1.17}{0.46} & \diff{1.38}{0.22} \\
    \hhline{|---||-||-|-|}
    CIFAR-100 & 100 & 50,000 & 35.87 $\pm$ 0.23 & \diff{3.99}{0.24} & \diff{3.31}{0.59} \\
    CIFAR-100 & 100 & 20,000 & 27.83 $\pm$ 0.34 & \diff{4.40}{0.32} & \diff{2.27}{0.37} \\
    CIFAR-100 & 100 & 10,000 & 21.96 $\pm$ 0.49 & \diff{2.70}{0.22} & \diff{2.67}{0.68} \\
    CIFAR-100 & 100 & {\leavevmode\hphantom{0}}5,000 & 17.20 $\pm$ 0.20 & \diff{2.35}{0.29} & \diff{1.92}{0.24} \\
    \hhline{|---||-||-|-|}
    Tiny-ImageNet & 200 &                         100,000 & 21.94 $\pm$ 0.19 & \diff{3.79}{0.34} & \diff{2.73}{0.49} \\
    Tiny-ImageNet & 200 & {\leavevmode\hphantom{0}}50,000 & 16.33 $\pm$ 0.32 & \diff{3.64}{0.47} & \diff{3.06}{0.33} \\
    Tiny-ImageNet & 200 & {\leavevmode\hphantom{0}}20,000 & 10.16 $\pm$ 0.22 & \diff{2.94}{0.36} & \diff{2.02}{0.34} \\
    Tiny-ImageNet & 200 & {\leavevmode\hphantom{0}}10,000 & {\leavevmode\hphantom{0}}7.38 $\pm$ 0.11 & \diff{2.01}{0.32} & \diff{1.14}{0.19} \\
    \hhline{|---||-||-|-|}
\end{tabularx}
\label{tab:results-single-label-staged}
\end{table*}